\documentclass[twoside]{article}

\usepackage[accepted]{aistats2024}
\RequirePackage[loading]{tracefnt}

\usepackage[utf8]{inputenc} 
\usepackage[T1]{fontenc}    
\usepackage{hyperref}       
\usepackage{url}            
\usepackage{booktabs}       
\usepackage{amsfonts}       
\usepackage{nicefrac}       
\usepackage{microtype}      
\usepackage{xcolor}         

\usepackage{amsfonts, amsmath, amssymb, amsthm}
\usepackage{xfrac}
\usepackage{graphicx, xcolor, colortbl}
\usepackage{footnote}
\usepackage{tablefootnote}
\usepackage{natbib}
\usepackage{dsfont}
\usepackage{bbm}
\usepackage{algorithm}
\usepackage{algorithmic}
\usepackage{import}
\usepackage{mathtools}
\usepackage{bm}
\usepackage{xspace}
\usepackage{thmtools,thm-restate} 
\usepackage{accents}
\usepackage{framed}
\usepackage[inline]{enumitem}
\usepackage{booktabs}
\usepackage{color, colortbl}
\usepackage{multirow}
\usepackage{multicol}
\usepackage{wrapfig}

\usepackage{notation}

\newtheorem{theorem}{Theorem}

\newtheorem{proposition}{Proposition}
\theoremstyle{definition}

\newtheorem{remark}{Remark}

\newtheorem*{proposition*}{Proposition}

\begin{document}

%

%
\runningauthor{Daniil Tiapkin$^\ast$, Nikita Morozov$^\ast$, Alexey Naumov, Dmitry Vetrov }

\twocolumn[

\aistatstitle{Generative Flow Networks as Entropy-Regularized RL}

\aistatsauthor{ Daniil Tiapkin$^\ast$ \And Nikita Morozov$^\ast$}

\aistatsaddress{  CMAP - CNRS -  \'Ecole polytechnique \\
-  Institut Polytechnique de
Paris,  \\ 
LMO, CNRS, Universit\'e Paris-Saclay, \\ HSE Unversity \And  HSE University} 

\aistatsauthor{Alexey Naumov \And Dmitry Vetrov }

\aistatsaddress{ HSE University, \\ Steklov Mathematical Institute \\
Russian Academy of Sciences  \And Constructor University, Bremen}]

\begin{abstract}
    The recently proposed generative flow networks (GFlowNets) are a method of training a policy to sample compositional discrete objects with probabilities proportional to a given reward via a sequence of actions. GFlowNets exploit the sequential nature of the problem, drawing parallels with reinforcement learning (RL). Our work extends the connection between RL and GFlowNets to a general case. We demonstrate how the task of learning a generative flow network can be efficiently redefined as an entropy-regularized RL problem with a specific reward and regularizer structure. Furthermore, we illustrate the practical efficiency of this reformulation by applying standard soft RL algorithms to GFlowNet training across several probabilistic modeling tasks. Contrary to previously reported results, we show that entropic RL approaches can be competitive against established GFlowNet training methods. This perspective opens a direct path for integrating RL principles into the realm of generative flow networks.
\end{abstract}

\section{INTRODUCTION}\label{sec:intro}

\vspace{-0.1cm}
Generative flow networks (GFlowNet, \citealp{bengio2021flow}) are models designed to learn a sampler from a complex discrete space according to a probability mass function given up to an unknown normalizing constant. A fundamental assumption underlying GFlowNets is an additional sequential structure of the space: any object can be generated by following a sequence of actions starting from an initial state, which induces a directed acyclic graph over "incomplete" objects. In essence, GFlowNet aims to learn a stochastic policy to select actions and navigate in this graph in order to match a distribution over final objects with the desired one. A motivational example of the work \cite{bengio2021flow} was molecule generation: GFlowNet starts from an empty molecule and learns a policy that adds various blocks sequentially in order to match a desired distribution over a certain molecule space.

The problem statement shares a lot of similarities with reinforcement learning (RL, \citealp{sutton2018reinforcement}) and indeed existing generative flow network algorithms often borrow ideas from RL. In particular, the original GFlowNet training method called flow matching \citep{bengio2021flow} was inspired by temporal-difference learning \citep{sutton1988learning}. Next, established techniques in RL were a large source of inspiration for developing novel GFlowNet techniques: subtrajectory balance loss \citep{madan2023learning} was inspired by $\mathrm{TD}(\lambda)$; a prioritized experience replay idea \citep{lin1992self,schaul2016prioritized} was actively used in \citep{shen2023towards,vemgal2023empirical}; distributional perspective on RL \citep{bellemare2023distributional} also appears in GFlowNet literature \citep{zhang2023distributional}. Additionally, various RL exploration techniques~\citep{osband2016deep,ostrovski2017count,burda2018exploration} were recently adapted into the GFlowNet framework~\citep{pan2023generative,rector2023thompson}.

This large source of techniques results in a question about a more rigorous connection between RL and GFlowNets. 
Unfortunately, RL with the classical goal of return maximization alone cannot adequately address the sampling problem since the final optimal policies are deterministic or close to deterministic. However, there exists a paradigm of entropy-regularized RL \citep{fox2016taming,haarnoja2017reinforcement,schulman2017equivalence} (also known as MaxEnt RL and soft RL), where a final goal is to find a policy that will not only maximize return but also be close to a uniform one, resulting in an energy-based representation of the optimal policies \citep{haarnoja2017reinforcement}.

The authors of \cite{bengio2021flow} have drawn a direct connection between MaxEnt RL and GFlowNets in the setup of autoregressive generation, or, equivalently, in the setting where the corresponding generation graph is a tree. However, the presented reduction does not hold in the setting of general graphs and results in generation from a highly biased distribution. After that, the authors of \cite{bengio2021gflownet} concluded that the direct connection holds \textit{only} in the tree setup.

\textit{In our work, we refute this claim and show that it is possible to reduce the GFlowNet learning problem to entropy-regularized RL beyond the setup of tree-structured graphs.}

The main corollary of our result is the possibility to directly apply existing RL techniques without the need to additionally adapt them within GFlowNet framework.

We highlight our main contributions:
\vspace{-0.25cm}
\begin{itemize}[itemsep=-2pt,leftmargin=8pt]
    \item We establish a connection between GFlowNets and entropy-regularized RL by direct reduction;
    \item We represent detailed balance \citep{bengio2021gflownet} and trajectory balance \citep{malkin2022trajectory} algorithms as existing soft RL algorithms;
    \item We show how to apply the proposed reduction practically by transmitting \SoftDQN \citep{haarnoja2017reinforcement} and \MunDQN \citep{vieillard2020munchausen} into the realm of GFlowNets;
    \item Finally, we provide experimental validation of our theoretical results and show that MaxEnt RL algorithms can be competitive or even outperform established GFlowNet training methods, contrary to some previously reported results \citep{bengio2021flow}. 
\end{itemize}
\vspace{-0.1cm}

Source code: \href{https://github.com/d-tiapkin/gflownet-rl}{github.com/d-tiapkin/gflownet-rl}.

\section{BACKGROUND}\label{sec:background}

\vspace{-0.1cm}

\subsection{Generative Flow Networks}

\vspace{-0.1cm}
In this section we provide some essential definitions and background on GFlowNets. All proofs of the following statements can be found in~\cite{bengio2021flow} and~\cite{bengio2021gflownet}.

Suppose we have a finite discrete space $\cX$ with a black-box non-negative reward function $\cR(x)$ given for every $x \in \cX$. Our goal is to construct and train a model that will sample objects from $\cX$ from the probability distribution $\cR(x) / \rmZ$ proportional to the reward, where $\rmZ$ is an unknown normalizing constant.


\vspace{-0.2cm}
\paragraph{Flows over Directed Acyclic Graphs} The generation process in GFlowNets is viewed as a sequence of constructive actions where we start with an "empty" object and at each step add some new component to it. To formally describe this process, a finite directed acyclic graph (DAG) $\cG = (\cS, \cE)$ is introduced, where $\cS$ is a state space and $\cE \subseteq \cS \times \cS$ is a set of edges. For each state $s \in \cS$ we define a set of actions $\cA_s$ as possible next states: $s' \in \cA_s \Leftrightarrow s \to s' \in \cE$. Each action corresponds to adding some new component to an object $s$, transforming it into a new object $s'$. If $s' \in \cA_s$ we say that $s'$ is a child of $s$ and $s$ is a parent of $s'$. There is a special initial state $s_0 \in \cS$ associated with an "empty" object, and it is the only state that has no incoming edges. All other states can be reached from $s_0$, and the set of terminal states (states with no outgoing edges) directly corresponds to the initial space of interest $\cX$. 




Denote $\cT$ as a set of all complete trajectories in the graph $\tau=\left(s_0 \to s_1 \to \ldots \to s_{n_{\tau}}\right)$, where $n_{\tau}$ is the trajectory length, $s_i \to s_{i + 1} \in \cE$, $s_0$ is always the initial state, and $s_{n_{\tau}}$ is always a terminal state, thus $s_{n_{\tau}} \in \cX$. As a result, any complete trajectory can be viewed as a sequence of actions that constructs the object corresponding to $s_{n_{\tau}}$ staring from $s_0$. A trajectory flow is any nonnegative function $\cF\colon \mathcal{T} \to \mathbb{R}_{\geq 0}$. If $\cF$ is not identically zero, it can be used to introduce a probability distribution over complete trajectories:
\vspace{-0.2cm}
\begin{normalsize}
\begin{equation*}
\cP(\tau) \triangleq \frac{1}{\rmZ_{\cF}} \cF(\tau), \quad \rm\rmZ_{\cF}\triangleq\sum_{\tau \in \mathcal{T}} \cF(\tau). 
\end{equation*}
\end{normalsize}
\vspace{-0.6cm}

After that, flows for states and edges are introduced as

\vspace{-0.7cm}
\begin{normalsize}
\begin{equation*}
\cF(s) \triangleq \sum_{\tau \ni s} \cF(\tau), \quad  \cF\left(s \to s^{\prime}\right) \triangleq \sum_{\tau \ni (s \to s') } \cF(\tau).
\end{equation*}
\end{normalsize}
\vspace{-0.6cm}

Then, {\normalsize $\cF(s) / \rm\rmZ_{\cF}$} corresponds to the probability that a random trajectory contains the state $s$, and {\normalsize $\cF(s \to s') / \rm\rmZ_{\cF}$} corresponds to the probability that a random trajectory contains the edge $s \to s'$. These definitions also imply {\normalsize $\rmZ_{\cF} = \cF(s_0) = \sum_{x \in \cX} \cF(x)$}.

Denote {\normalsize $\cP(x)$} as the probability that $x$ is the terminal state of a trajectory sampled from {\normalsize$\cP(\tau)$}. If \textit{the reward matching constraint} is satisfied for all terminal states $x \in \cX$:
\begin{normalsize}
\begin{equation}\label{eq:reward_matching}
\cF(x) = \cR(x),
\end{equation}
\end{normalsize}

{\normalsize$\cP(x)$} will be equal to {\normalsize$\cR(x) / \rmZ$}. Thus by sampling trajectories we can sample objects from the distribution of interest. 


\vspace{-0.2cm}
\paragraph{Markovian Flows} To allow efficient sampling of trajectories, the family of trajectory flows considered in practice is narrowed down to Markovian flows. A trajectory flow $\cF$ is said to be Markovian if $\forall s \in \cS \setminus \cX$ there exist distributions $\PF(-|s)$ over the children of $s$ such that $\cP$ can be factorized as 
\vspace{-0.05cm}
\begin{normalsize}
\begin{equation*}
\cP\left(\tau=\left(s_0 \rightarrow \ldots \rightarrow s_{n_\tau}\right)\right)=\prod_{t=1}^{n_\tau} \PF\left(s_t | s_{t-1}\right).
\end{equation*}
\end{normalsize}
\vspace{-0.05cm}
\!\!This means that trajectories can be sampled by iteratively sampling next states from $\PF(- \mid s_i)$ (sequential construction of an object). Equivalently, $\cF$ is Markovian if $\forall s \in \cS \setminus \{ s_0 \}$ there exist distributions $\PB(-|s)$ over the parents of $s$ such that $\forall x \in \cX$
\vspace{-0.05cm}
\begin{equation*}
\cP\left(\tau=\left(s_0 \rightarrow \ldots \rightarrow s_{n_\tau}\right) | s_{n_\tau} = x\right)=\prod_{t=1}^{n_\tau} \PB\left(s_{t-1} | s_{t}\right),
\end{equation*}
\vspace{-0.05cm}
\!\!\!which means that trajectories ending at $x$ can be sampled by iteratively sampling previous states from $\PB(- \mid s_i)$. If $\cF$ is Markovian, then $\PF$ and $\PB$ can be uniquely defined as 
\begin{normalsize}
\begin{equation}\label{eq:pf_pb_definition}
\PF\left(s^{\prime} | s\right)=\frac{\cF\left(s \rightarrow s^{\prime}\right)}{\cF(s)}, \quad \PB\left(s | s^{\prime}\right)=\frac{\cF\left(s \rightarrow s^{\prime}\right)}{\cF\left(s^{\prime}\right)}.
\end{equation}
\end{normalsize}

\vspace{-0.2cm}
We call $\PF$ and $\PB$ the forward policy and the backward policy respectively. \textit{The detailed balance constraint} is a relation between them:
\begin{normalsize}
\begin{equation}\label{eq:detailed_balance}
\cF(s)\PF(s' | s) = \cF(s')\PB(s | s').
\end{equation}
\end{normalsize}

\vspace{-0.2cm}
\textit{The trajectory balance constraint} states that for any complete trajectory
\vspace{-0.2cm}
\begin{normalsize}
\begin{equation}\label{eq:trajectory_balance}
\prod_{t=1}^{n_\tau} \PF\left(s_t | s_{t-1}\right) = \frac{\cF(s_{n_\tau})}{\rmZ_{\cF}}\prod_{t=1}^{n_\tau} \PB\left(s_{t-1} | s_{t}\right).
\end{equation}
\end{normalsize}

\vspace{-0.2cm}
Thus any Markovian flow can be uniquely determined from $Z_{\cF}$ and $\PF(- | s) \; \forall s \in \cS \setminus \cX$ or from $\cF(x) \; \forall x \in \cX$ and $\PB(-|s) \; \forall s \in \cS \setminus \{ s_0 \}$.


\vspace{-0.1cm}
\paragraph{Training GFlowNets} In essence, GFlowNet is a model that parameterizes Markovian flows over some fixed DAG and is trained to minimize some objective. The objective should be chosen in such a way that, if the model is capable of parameterizing any Markovian flow, global minimization of the objective will lead to producing a flow that satisfies~\eqref{eq:reward_matching}. Thus the forward policy of a trained GFlowNet can be used as a sampler from the reward distribution. Two of the widely used training objectives are:


\textit{Trajectory Balance} (\TB). Introduced in~\cite{malkin2022trajectory}, based on the trajectory balance constraint~\eqref{eq:trajectory_balance}. A model with parameters $\theta$ predicts the normalizing constant $\rmZ_{\theta}$, forward policy $\PF(-| s, \theta)$ and backward policy $\PB(-| s, \theta)$. The trajectory balance objective is defined for each complete trajectory $\tau=\left(s_0 \to s_1 \to \ldots \to s_{n_\tau} = x\right)$:
\begin{normalsize}
\begin{equation}\label{eq:TB_loss}
\mathcal{L}_{\mathrm{TB}}(\tau) = \left(\log \frac{\rmZ_{\theta} \prod_{t=1}^{n_\tau} \PF(s_t | s_{t - 1}, \theta)}{\cR(x) \prod_{t=1}^{n_\tau} \PB(s_{t - 1} | s_{t}, \theta)} \right)^2.
\end{equation}
\end{normalsize}


\textit{Detailed Balance} (\DB). Introduced in~\cite{bengio2021gflownet}, based on the detailed balance constraint~\eqref{eq:detailed_balance}. A model with parameters $\theta$ predicts state flows $\cF_\theta(s)$, forward policy $\PF(- | s, \theta)$ and backward policy $\PB(- | s, \theta)$. It is important to note that such parameterization is excessive and the components are not necessarily compatible with each other during the course of training. The detailed balance objective is defined for each edge $s \to s' \in \cE$:
\begin{normalsize}
\begin{equation}\label{eq:DB_loss}
\mathcal{L}_{\mathrm{DB}}(s, s^{\prime}) = \left(\log \frac{\cF_{\theta}(s) \PF(s' | s, \theta)}{\cF_{\theta}(s')\PB(s | s', \theta)} \right)^2.
\end{equation}
\end{normalsize}
\!\!For terminal states $\cF_{\theta}(x)$ is substituted with $\cR(x)$ in the objective. 

During the training, the objective of choice is stochastically optimized across complete trajectories sampled from a training policy (which is usually taken to be $\PF$, a tempered version of it or a mixture of $\PF$ and a uniform distribution).  An important note is that for all objectives $\PB(- \mid s)$ can either be trained or fixed, e.g. to be uniform over the parents of $s$. When the backward policy is fixed, the corresponding forward policy is uniquely determined by $\PB$ and $\cR$.

Among other existing objectives, the originally proposed flow matching objective~\cite{bengio2021flow} was shown to have worse performance than \TB and \DB~\citep{malkin2022trajectory, madan2023learning}. Recently, \textit{subtrajectory balance} (\SubTB)~\citep{madan2023learning} was introduced as a generalization of \TB and \DB and shown to have better performance when appropriately tuned.

\subsection{Soft Reinforcement Learning}\label{sec:soft_rl_background}

In this section we describe the main aspects of usual reinforcement learning and its entropy-regularized version, also known as MaxEnt and Soft RL.

\vspace{-0.3cm}
\paragraph{Markov Decision Process} The main theoretical object of RL is Markov decision process (MDP) that is defined as a tuple $\cM = (\cS, \cA, \MK, r, \gamma, s_0)$, where $\cS$ and $\cA$ are finite state and action spaces, $\MK$ is a Markovian transition kernel, $r$ is a bounded deterministic reward function, $\gamma \in [0,1]$ is a discount factor, and $s_0$ is a fixed initial state. Additionally, for each state $s$ we define a set of possible actions as $\cA_s \subseteq \cA$. An RL agent interacts with MDP by a policy $\pi$ that maps each state $s$ to a probability distribution over possible actions.

\paragraph{Entropy-Regularized RL}
The performance measure of the agent in classic RL is a value that is defined as an expected (discounted) sum of rewards over a trajectory generated by following a policy $\pi$ starting from a state $s$. In contrast, in entropy-regularized reinforcement learning \citep{neu2017unified,geist2019theory} the value is augmented by Shannon entropy: 
\begin{equation}\label{eq:regularized_value_def}
    \textstyle V^\pi_{\lambda}(s) \triangleq \E_{\pi}\bigg[ \sum\limits_{t=0}^\infty \gamma^t (r(s_t,a_t) + \lambda \cH(\pi(s_t))) | s_0 = s\bigg],
\end{equation}
where $\lambda$ is a regularization coefficient, $\cH$ is a Shannon entropy function, $a_t \sim \pi(s_t), s_{t+1} \sim \MK(s_t,a_t)$ for all $t \geq 0$, and expectation is taken with respect to these random variables. In a similar manner we can define regularized Q-values  $Q^\pi_{\lambda}(s,a)$ as an expected (discounted) sum of rewards augmented by Shannon entropy  given a fixed initial state $s_0=s$ and an action $a_0 = a$. A regularized optimal  policy $\pistar_{\lambda}$ is a policy that maximizes $V^{\pi}_{\lambda}(s)$ for any initial state $s$. 

\vspace{-0.27cm}
\paragraph{Soft Bellman Equations} 
Let $\Vstar_{\lambda}$ and $\Qstar_{\lambda}$ be the value and the Q-value of the optimal policy $\pistar_{\lambda}$ correspondingly. Then Theorem~1 and 2 by \cite{haarnoja2017reinforcement} imply the following system relations for any $s \in \cS, a \in \cA_s$
\begin{normalsize}
\begin{align}\label{eq:soft_bellman_equations}
    \begin{split}
        \Qstar_{\lambda}(s,a) &\triangleq r(s,a) + \gamma \E_{s' \sim \MK(s,a)}[\Vstar(s')] \\
        \Vstar_{\lambda}(s) &\triangleq \logsumexp_{\lambda}\left( \Qstar_{\lambda}(s, \cdot)\right),
    \end{split}
\end{align}
\end{normalsize}

\vspace{-0.3cm}
where $\logsumexp_{\lambda}(x) \triangleq \lambda \log(\sum_{i=1}^d \rme^{x_i / \lambda})$, 
and the optimal policy is defined as a $\lambda$-softmax policy with respect to optimal Q-values $\pistar_{\lambda}(s) = \softmax_{\lambda}(\Qstar_{\lambda}(s, \cdot))$, where $[\softmax_{\lambda}(x)]_i \triangleq \rme^{x_i /\lambda}/(\sum_{j=1}^d \rme^{x_j / \lambda} )$, or, alternatively
\vspace{-0.25cm}
\begin{normalsize}
\begin{equation}\label{eq:soft_optimal_policy}
    \textstyle \pistar_{\lambda}(a|s) = \exp\left( \frac{1}{\lambda}\left( \Qstar_{\lambda}(s,a) - \Vstar_{\lambda}(s) \right) \right).
\end{equation}
\end{normalsize}

\vspace{-0.25cm}
Notice that as $\lambda \to 0$ we recover the well-known Bellman equations (see e.g. \cite{sutton2018reinforcement}) since $\logsumexp_{\lambda}$ approximates maximum and $\softmax_{\lambda}$ approximates a uniform distribution over $\argmax$.

\vspace{-0.27cm}
\paragraph{Soft Deep Q-Network} One of the classic ways to solve RL problem for discrete action spaces is the well-known Deep Q-Network algorithm (\DQN) \citep{mnih2015human}.  Below we present the generalization of this algorithm to entropy-regularized RL problems called \SoftDQN (also known as \SoftQLearning \citep{fox2016taming,haarnoja2017reinforcement,schulman2017equivalence} in a continuous control setting).

The goal of \SoftDQN algorithm is to approximate a solution to soft Bellman equations \eqref{eq:soft_bellman_equations} using a Q-value parameterized by a neural network. Let $Q_{\theta}$ be an online Q-network that represents the current approximation of the optimal regularized Q-value. Also let $\pi_{\theta} = \softmax_{\lambda}(Q_\theta)$ be a policy that approximates the optimal policy. Then, akin to \DQN, agent interacts with MDP using policy $\pi_{\theta}$ (with an additional $\varepsilon$-greedy exploration or even without it) and collects transitions $(s_t, a_t, r_t, s_{t+1})$ for $r_t = r(s_t,a_t)$ in a replay buffer $\cB$.  Then \SoftDQN performs stochastic gradient descent on the regression loss {\normalsize$\cL_{\SoftDQN}(\cB) = \E_{\cB}[(Q_{\theta}(s_t,a_t) - y_{\SoftDQN}(r_t, s_{t+1}) )^2]$} with the target Q-value
\begin{normalsize}
\begin{equation}\label{eq:soft_dqn_loss}
    y_{\SoftDQN}(r_t, s_{t+1}) = r_t + \gamma\logsumexp_{\lambda}\left( Q_{\bar \theta}(s_{t+1}, \cdot) \right),
\end{equation}
\end{normalsize}

\vspace{-0.1cm}
where $Q_{\bar \theta}$ is a target Q-network with parameters $\bar \theta$ that are periodically copied from the parameters of online Q-network $\theta$. Additionally, there exist other algorithms that were generalized from traditional RL to soft RL, such as celebrated \SAC \citep{haarnoja2018soft} and its discrete version \citep{christodoulou2019soft}. 

It is worth mentioning another algorithm such as Munchausen DQN (\MDQN, \citealp{vieillard2020munchausen}) that exploits the regularization structure similarly to \SoftDQN but augments the target using the current policy:
\[
    y_{\MDQN}(s_t, a_t, r_t, s_{t+1}) = y_{\SoftDQN}(r_t, s_{t+1}) + \alpha \lambda \log \pi_{\bar \theta}(a_t | s_t).
\]
It turns out that this scheme is equivalent (up to a reparametrization) to approximate mirror descent value iteration \citep{vieillard2020leverage} method for solving $(1-\alpha)\lambda$ entropy-regularized MDP with an additional KL-regularization with respect to the target policy with coefficient $\alpha \lambda$ \citep[Theorem~1]{vieillard2020munchausen}.
\section{REPRESENTATION OF GFLOWNETS AS SOFT RL}\label{sec:representation}

In this section we describe a deep connection between GFlowNets and entropy-regularized reinforcement learning.

\subsection{Prior Work}

The sequential type of the problem that GFlowNet is solving creates the parallel automatically, and the main inspiration for flow-matching loss arises from temporal-difference learning \citep{bengio2021flow}. 

Beyond inspiration, one formal link between GFlowNets and RL was established in the seminal paper \citep{bengio2021flow}. In particular, the authors showed that it is possible to learn the GFlowNet forward policy by applying MaxEnt RL in the setup of tree DAG. However, for a general setting the authors claimed that MaxEnt RL is sampling with probability proportional to $n(x) \cR(x)$ where $n(x)$ is a number of paths to a terminal state $x \in \cX$, thus these methods should be highly biased. Therefore, the authors of \citep{bengio2021gflownet} claimed that equivalence between MaxEnt RL and GFlowNet holds only if the underlying DAG has a tree structure.


In addition, it is possible to treat the GFlowNet learning problem as a convex MDP problem \citep{zahavy2021reward} with a goal to directly minimize $\KL(d^{\pi} \Vert \cR / \rmZ)$, where $d^\pi$ is a distribution induced by a policy $\pi$ over terminal states. However, as it was claimed by \cite{bengio2021gflownet}, RL-based algorithms for solving this problem require high-quality estimation of a distribution $d^\pi$, and the scheme by \cite{zahavy2021reward} needs to do it for changing policies during the learning process. A similar problem arises in the generalization of count-based exploration in deep RL \citep{bellemare2016unifying,ostrovski2017count,saade2023unlocking}.

\subsection{Direct Reduction by MDP Construction} 

In this section we show how to recast the problem of learning GFlowNet forward policy $\PF$ given a reward function $\cR$ and a backward policy $\PB$ as a soft RL problem.

First, for any DAG $\cG = (\cS, \cE)$ with a set of terminal states $\cX$ we can define the corresponding deterministic controlled Markov process $(\cS', \cA, \MK)$ with an additional absorbing state $s_f$. Formally, the construction is as follows. Firstly, choose the state space $\cS'$ as the state space of the corresponding DAG $\cG$ with an additional absorbing state $s_f$: $\cS' = \cS \cup \{s_f\}$. Next, instead of defining the full action space $\cA$, for any $s$ we select $\cA_s \subseteq \cS$ as a set of actions in state $s$ in $\cS \setminus \cX$, and define $\cA_{x} \triangleq \{s_f\}$ for any $x\in \cX \cup \{s_f\}$. In other words, each terminal state only has an action that leads to $s_f$ and $s_f$ itself has a loop. Then we can define $\cA$ as a union of all $\cA_s$ for any $s \in \cS'$. Finally, the construction of a deterministic Markov kernel is simple: $\MK(s'|s,a) = \ind\{s' = a\}$ since $a$ corresponds to the next state in $\cS$ given the corresponding action. Subsequently, we will write $s'$ instead of $a$ to emphasize this connection.

\begin{theorem}\label{th:gflownet_reduction}
    Let $\cG = (\cS, \cE)$ be a DAG with a set of terminal states $\cX$, let $\PB$ be a fixed backward policy and $\cR$ be a GFlowNet reward function.
    
    Let $\cM_{\cG} = (\cS', \cA, \MK, r, \gamma,s_0)$ be an MDP with $\cS', \cA, \MK$ constructed from $\cG$ with an additional absorbing state $s_f$, $\gamma=1$, and MDP rewards defined as follows
    \begin{normalsize}
    \begin{equation}\label{eq:def_mdp_reward}
        r(s,s') = \begin{cases}
            \log \PB(s| s') & s \not \in \cX \cup \{s_f\}, \\
            \log \cR(s) & s \in \cX, \\
            0 & s = s_f
        \end{cases}
    \end{equation}
    \end{normalsize}
    
    \vspace{-0.3cm}
    Then the optimal policy $\pistar_1(s'|s)$ for the regularized MDP with coefficient $\lambda=1$ is equal to $\PF(s'|s)$ for all $s \in \cS \setminus \cX, s' \in \cA_s$; 
    
    Moreover, $\forall s \neq s_f, s' \neq s_f$ regularized optimal value $\Vstar_1(s)$ and Q-value $\Qstar_1(s,s')$ coincide with $\log \cF(s)$ and $\log \cF(s \to s')$ respectively, where $\cF$ is the Markovian flow defined by $\PB$ and the GFlowNet reward $\cR$.
\end{theorem}

The complete proof is given in Appendix~\ref{app:theory} and basically uses soft Bellman equations \eqref{eq:soft_bellman_equations} and backward induction over a topological ordering of the graph.

\textbf{Intuition.} Given a fixed GFlowNet backward
policy, the optimal policy of MaxEnt RL with $\lambda=1$ and adjusted rewards is identical to the GFlowNet forward policy. The key insight lies in introducing intermediate rewards through the fixed backward policy in order to normalize probabilities of different paths leading to the same terminal state.

\begin{remark}
    In the setting of tree-structures DAG $\cG$ we notice that our result shows that it is enough to define the reward $\log \cR$ for terminal states since any backward policy $\PB$ satifies $\log \PB(s|s') = 0$. Thus, we reproduce the known result of Proposition~1(a,b) by \cite{bengio2021flow}, see also \citep{bengio2021gflownet}. Regarding the results of a part (c), the key difference between our construction of MDP and construction by \cite{bengio2021flow} is non-trivial rewards for non-terminal states that "compensate" a multiplier $n(x)$ that appears in \citep{bengio2021flow}. 
\end{remark}

\begin{remark}
Notably, our derivation explains a logarithmic scaling that appears in \TB (see \eqref{eq:TB_loss}), \DB (see \eqref{eq:DB_loss}) and \SubTB~\citep{madan2023learning} losses. Previously, the appearance of these logarithms was explained solely by numerical reasons \citep{bengio2021flow,bengio2021gflownet}.
\end{remark}
\begin{remark}
    Taking $\lambda\not=1$ under the choice of rewards established in Theorem~\ref{th:gflownet_reduction} will lead to a biased policy. To see it, notice that solving soft RL with coefficient $\lambda>0$ and rewards $r$ is equivalent to solving soft RL with coefficient $\lambda' =1$ and new rewards $r' = r/\lambda$ since $V_{\lambda}^\star(s_0; r) = \mathbb{E}[\sum_t r_t + \lambda \mathcal{H}(\pi(s_t))] = \lambda \mathbb{E}[\sum_t r_t/\lambda +  \mathcal{H}(\pi(s_t))] = \lambda V_{1}^\star(s_0; r/\lambda)$. For the terminal rewards it will lead to rescaling the GFlowNet reward $R(x) \mapsto R^{1/\lambda}(x)$, but all intermediate rewards will also be rescaled $\log P_{B}(s|s') \mapsto \log P_{B}(s|s') / \lambda$, leading to a bias. However, additionally multiplying intermediate rewards by $\lambda$ in Theorem~\ref{th:gflownet_reduction} for $\lambda \neq 1$ will lead to a policy that samples from the tempered GFlowNet reward distribution.
\end{remark}
\begin{remark}
    We highlight two major differences with the experimental setup of \cite{bengio2021flow}: they applied PPO  \citep{schulman2017proximal} with regularization coefficient $\lambda < 1$ and \SAC \citep{haarnoja2018soft} with adaptive regularization coefficient (whilst our construction requires $\lambda = 1$), and set intermediate rewards to zero (while we set them to $\log \PB(s \mid s')$) This resulted in convergence to a biased final policy.
\end{remark}

\vspace{-0.25cm}
\paragraph{Interpretation of Value}

Notably, in the MDP $\cM_{\cG}$ we have a clear interpretation of the value function. 
But first we need to introduce several important definitions. For a policy $\pi$ in a deterministic MDP $\cM_{\cG}$ with all trajectories of length $n_{\tau} \leq N$ we can define an induced trajectory distribution as follows
\vspace{-0.15cm}
\begin{normalsize}
\[
    \textstyle q^{\pi}(\tau) = \prod_{i=1}^{N} \pi(s_{i} | s_{i-1}) = \prod_{i=1}^{n_{\tau}} \pi(s_{i} | s_{i-1}),
\]
\end{normalsize}

\vspace{-0.2cm}
where $\tau = (s_0 \to \ldots \to s_N)$ with $s_N = s_f$ and $s_{n_\tau} \in \cX$. The last identity holds since $\pi(s_f| s_f) = \pi(s_f | x) = 1$ for any $x \in \cX$. Additionally, we define a marginal distribution of sampling by policy $\pi$ as $d^{\pi}(x) = \P_{\pi}\left[ s_{n_{\tau}} = x \right]$. Additionally, for a fixed GFlowNet reward $\cR$ and a backward policy $\PB$ we define a corresponding distribution over trajectories with a slight abuse of notation
\vspace{-0.15cm}
\begin{normalsize}
\[
    \PB(\tau) = \frac{\cR(s_{n_{\tau}})}{\rmZ} \prod_{i=1}^{n_{\tau}} \PB(s_{i-1} | s_i).
\]
\end{normalsize}

\vspace{-0.2cm}
Notice that by the trajectory balance constraint \eqref{eq:trajectory_balance} and by Theorem~\ref{th:gflownet_reduction} $\PB(\tau)$ is equal to $q^{\pistar_1}(\tau)$.

\begin{proposition}\label{prop:value_expression}
    Let $\cM_{\cG}$ be an MDP constructed in Theorem~\ref{th:gflownet_reduction} by a DAG $\cG$ given a fixed backward policy $\PB$ and a GFlowNet reward function $\cR$. Then for any policy $\pi$ its regularized value $V^{\pi}_1$ in the initial state $s_0$ is equal to 
    $V^{\pi}_1(s_0) =\log \rmZ - \KL(q^{\pi} \Vert \PB).$
\end{proposition}
\vspace{-0.15cm}

The proof idea is to apply the direct definition of value \eqref{eq:regularized_value_def}. The complete proof is given in Appendix~\ref{app:theory}.

The main takeaway of this proposition is that maximization of the value function in the prescribed MDP $\cM$ is equivalent to minimization of the KL-divergence between trajectory distribution induced by the current policy and by the backward policy. Moreover, it shows that all near-optimal policies in soft RL-sense introduce a flow with close trajectory distributions, and, moreover, close marginal distribution:
\vspace{-0.1cm}
\begin{normalsize}
\[
    V^\star_1(s_0) - V^{\pi}_1(s_0) = \KL(q^{\pi} \Vert \PB) \geq \KL(d^{\pi} \Vert \cR / \rmZ ),
\]
\end{normalsize}

\vspace{-0.15cm}
where $d^\pi(x)$ is a marginal distribution over terminal states induced by $\pi$ and $\cR(x) / \rmZ$ is the reward distribution we want to sample from, see \cite[Theorem 2.5.3]{cover1999elements}. In other words, the policy error controls the distribution approximation error, additionally justifying the presented MDP construction.

\vspace{-0.1cm}
\subsection{Soft DQN in the realm of GFlowNets}

\vspace{-0.1cm}
In this section we explain how to apply \SoftDQN algorithm to GFlowNet training.

As described in Section~\ref{sec:soft_rl_background}, in \SoftDQN an agent uses a network $Q_{\theta}$ that approximates optimal $Q$-value or, equivalently, the logarithm of the edge flow. To interact with DAG the agent uses a softmax policy $\pi_\theta(s) = \softmax_1(Q_{\theta}(s,\cdot))$. 
During the interaction, the agent collects transitions $(s_t, a_t, r_t, s_{t+1})$ into a replay buffer $\cB$ and optimizes a regression loss $\cL_{\SoftDQN}$ using targets \eqref{eq:soft_dqn_loss} with rewards $r_t = \log \PB(s_{t} | s_{t+1})$ for any $s_t \not \in \cX$. For terminal transitions with $s_t \in \cX$ and $s_{t+1} = s_f$ the target is equal to $ \log \cR(s_t)$. In particular, using the interpretation $Q_{\theta}(s,s') = \log \cF_{\theta}(s\to s')$ we can obtain the following loss
\begin{equation}\label{eq:soft_dqn_loss_flow}
    \cL_{\SoftDQN}(\cB) = \E_{\cB}\left[\left( \log \frac{\cF_{\theta}(s_t\to s_{t+1})}{\PB(s_{t}| s_{t+1}) \cF_{\bar \theta}(s_{t+1}) } \right)^2\right],
\end{equation}

where $\cF_{\bar\theta}(x) = \cR(x) \; \forall  x \in \cX$, and otherwise {\normalsize$\cF_{\bar\theta}(s) \triangleq \sum_{s'} \cF_{\bar\theta}(s\to s')$}. $\bar\theta$ are parameters of a target network that is updated with weights $\theta$ from time to time. In particular, this loss may be "derived" from the definition of backward policy \eqref{eq:pf_pb_definition} akin to deriving \DB and \TB losses from the corresponding identities \eqref{eq:detailed_balance}-\eqref{eq:trajectory_balance} with the differences of fixed backward policy and usage of a target network $\bar \theta$ and a replay buffer $\cB$. We refer to Appendix~\ref{app:algo} for discussion on technical algorithmic choices.

\vspace{-0.3cm}
\paragraph{Munchausen DQN}
Using exactly the same representation, we can apply \MDQN~\citep{vieillard2020munchausen} to train GFlowNets. However, unlike \SoftDQN, it requires a more careful choice of $\lambda$ since \MDQN solves a regularized problem with entropy coefficient $(1-\alpha) \lambda$. Thus, one needs to opt for a non-trivial choice of $\lambda = 1/(1-\alpha)$ to compensate the effect of Munchausen penalty. We refer the reader to Appendix~\ref{app:algo} for a more detailed description.

\vspace{-0.3cm}
\paragraph{Learnable backward policy} The aforementioned approach works with fixed backward policies, which is a viable and widely utilized option for GFlowNet training~\citep{malkin2022trajectory, zhang2022generative, deleu2022bayesian, malkin2023gflownets, madan2023learning} that has benefits such as more stable training. At the same time having a trainable backward policy can lead to faster convergence~\citep{malkin2022trajectory}. Our approach still lacks understanding of the effect of optimization of the backward policy, which is a limitation.

Nevertheless, it seems to be possible to interpret this optimization in a game-theoretical fashion: one player is selecting rewards (defined by a backward policy) to optimize its objective and the other player is trying to guess a good forward policy for these rewards, see \citep{zahavy2021reward,tiapkin2023fast} for a similar idea applied to convex MDPs and maximum entropy exploration. We leave the formal derivation as a promising direction for future research. 

\subsection{Existing GFlowNets through Soft RL}\label{sec:existing_gflow_as_softrl}

In this section we show that \DB and \TB GFlowNet losses can be derived from existing RL literature up to technical details.

\vspace{-0.3cm}
\paragraph{Detailed Balance as Dueling Soft DQN} 
To show a more direct connection between \SoftDQN and \DB algorithms, we first describe a well-known technique in RL known as a \textit{dueling architecture} \citep{wang2016dueling}. Instead of parameterizing Q-values solely, the dueling architecture proposes to have a neural network with two streams for values {\normalsize $V_{\theta}(s)$} and action advantages {\normalsize $A_{\theta}(s,a)$}. These two streams are connected to the original Q-value as $Q_{\theta}(s,a) = V_{\theta}(s) + A_{\theta}(s,a)$ by the definition of advantage function.

The main idea behind this technique is that it is believed that value could be learnt faster and then accelerate the learning of advantage by bootstrapping \citep{tang2023va}. However, as it was acknowledged in the seminal work \citep{wang2016dueling}, the reparametrization presented above is unidentifiable since the advantage is not uniquely defined by a Q-value. In the setup of classic RL it was suggested to subtract maximum of the advantage. In the setup of soft RL it makes sense to subtract log-sum-exp of the advantage and obtain the following identity
\vspace{-0.1cm}
\begin{normalsize}
\begin{align*}
    Q_{\theta}(s,a) = V_{\theta}(s) + A_{\theta}(s,a) - \logsumexp_1(A_{\theta}(s,\cdot)).
\end{align*}
\end{normalsize}

\vspace{-0.3cm}
The key properties of this representation are the following relations (for $\lambda = 1$ in the GFlowNet setup):
\vspace{-0.15cm}
\begin{normalsize}
\begin{align*}
\begin{split}
    \logsumexp_1 (Q_{\theta}(s,\cdot)) &= V_{\theta}(s), \\ \log \pi_{\theta}(a|s) = A_{\theta}(s,a) &- \logsumexp_1(A_{\theta}(s,\cdot)).
\end{split}
\end{align*}
\end{normalsize}

\vspace{-0.25cm}
Using GFlowNet interpretation of {\normalsize $V_{\theta}(s) = \log \cF(s)$} and {\normalsize $\pi_{\theta}(s'| s) = \PF(s' | s, \theta)$}, the loss \eqref{eq:soft_dqn_loss_flow} transforms into 
\begin{normalsize}
\[
    \cL^{\mathrm{Dueling}}_{\SoftDQN}(\cB) = \E_{\cB}\left[ \left( \log \frac{\PF(s_{t+1} | s_t, \theta) \cF_{\theta}(s_t))}{\PB(s_t | s_{t+1}) \cF_{\bar \theta}(s_{t+1}) } \right)^2\right],
\]
\end{normalsize}

\vspace{-0.35cm}
that coincides with \DB-loss \eqref{eq:DB_loss} up to the lack of a learnable backward policy and usage of a replay buffer and a target network. 

\vspace{-0.3cm}
\paragraph{Trajectory Balance as Policy Gradient}

Let us consider the setting of policy gradient methods: we parameterize the policy space $\pi_{\theta}$ by parameter $\theta$ and our goal is to directly maximize $V^{\pi_{\theta}}(s_0)$. The essence of policy gradient methods is to apply stochastic gradient ascent over $\theta$ to optimize this objective. 

However, a combination of Proposition~\ref{prop:value_expression} and Proposition~1 by \citep{malkin2023gflownets} implies that
\vspace{-0.1cm}
\begin{normalsize}
\[
   \textstyle -\nabla_\theta V^{\pi_{\theta}}_1(s_0) = \nabla_{\theta} \KL(q^{\pi_\theta}\Vert \PB) = \frac{1}{2} \E_{\tau \sim \pi_\theta}\left[ \nabla_{\theta} \cL_{\mathrm{TB}}(\tau) \right],
\]
\end{normalsize}

\vspace{-0.15cm}
where $\cL_{\mathrm{TB}}(\tau)$ is $\TB$-loss \eqref{eq:TB_loss}. This means that optimization of \textit{on-policy} \TB-loss is equivalent to policy gradient with using $\log \rmZ_{\theta}$ as a baseline function.  Additionally, this connection is very close to a connection between soft Q-learning and policy gradient methods established by \cite{schulman2017equivalence}.


\section{EXPERIMENTS}\label{sec:experiments}

\begin{figure*}[!t]

    \centering
    \includegraphics[width=0.2599\linewidth]{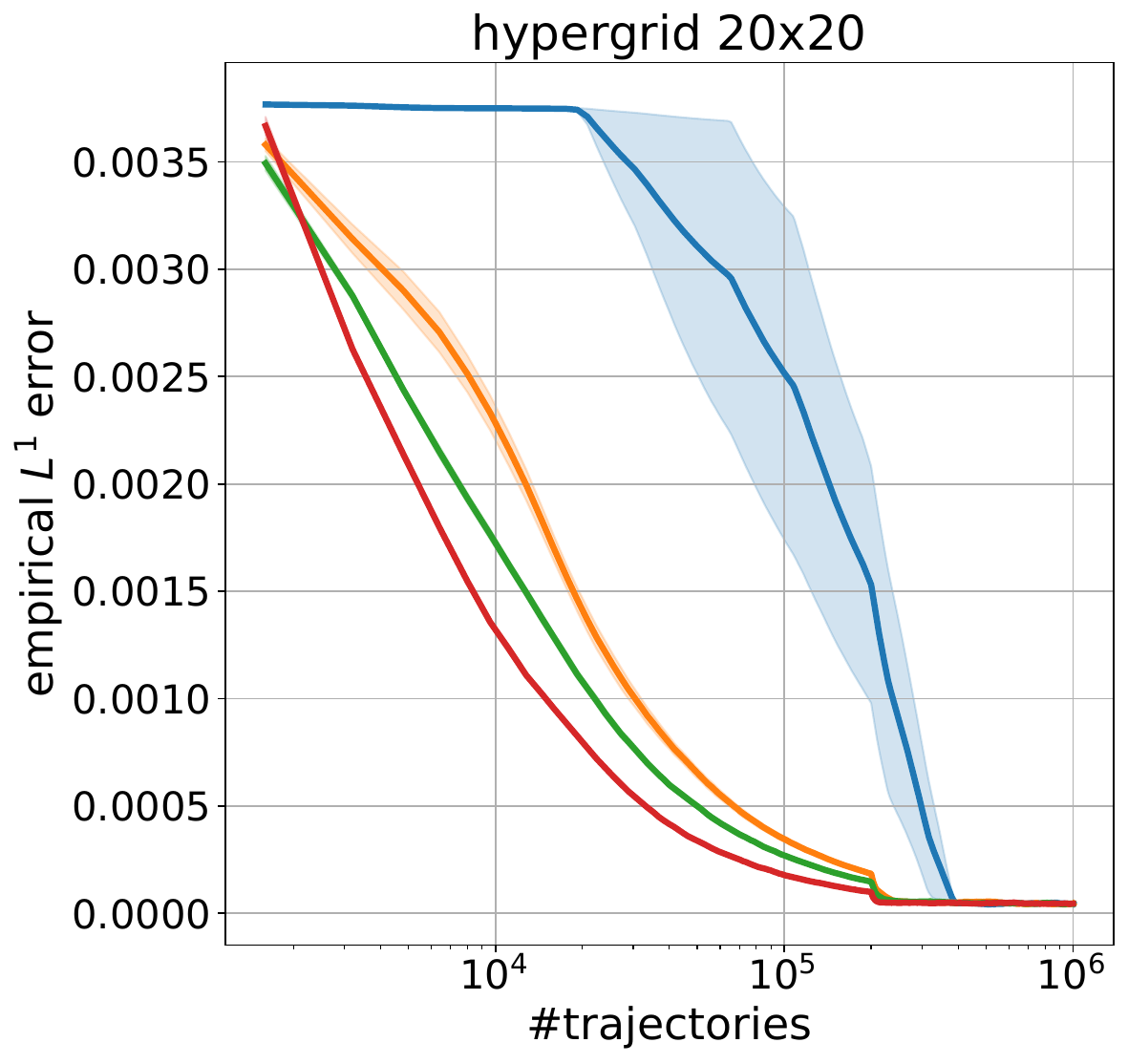}
    \includegraphics[width=0.2290\linewidth]{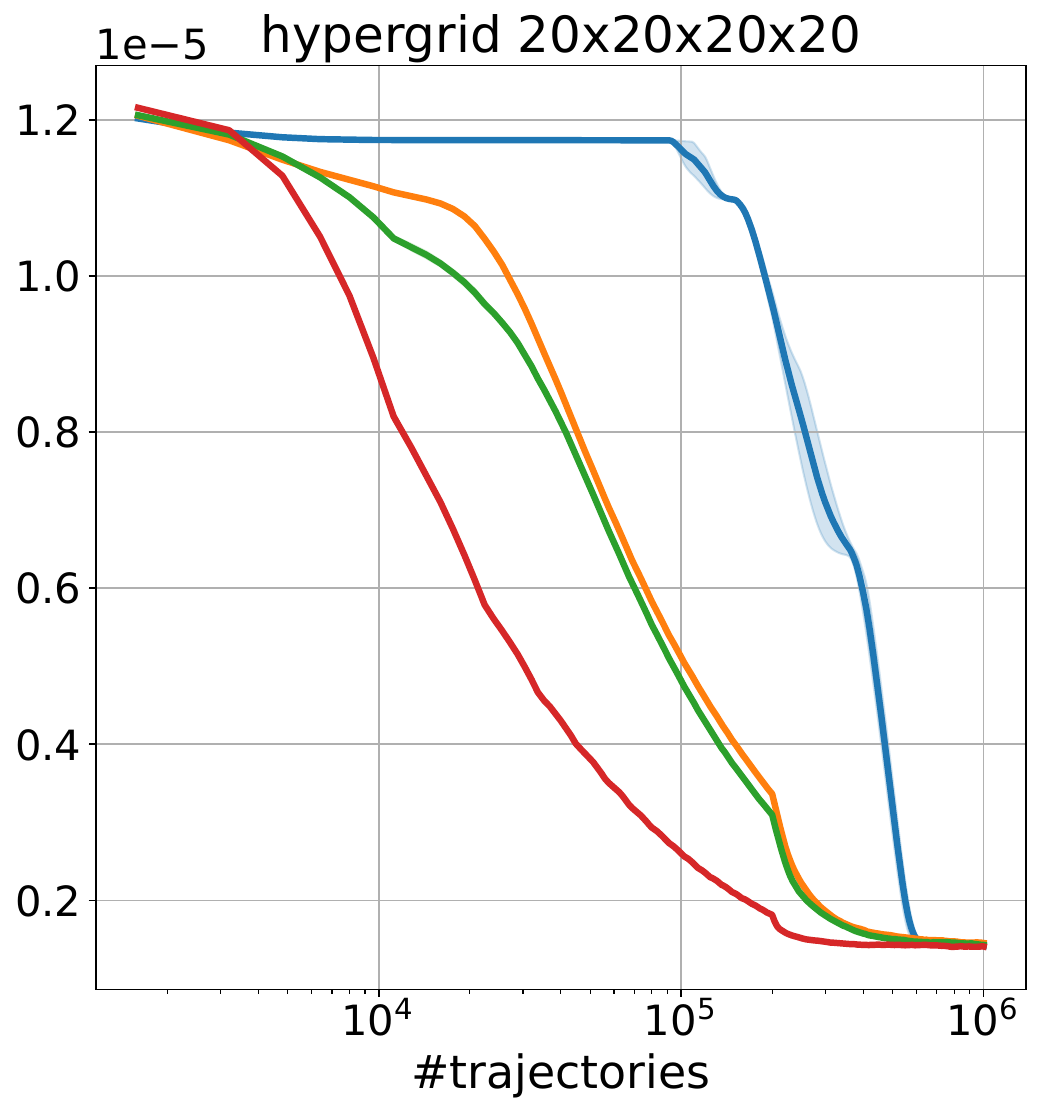}
    \includegraphics[width=0.2508\linewidth]{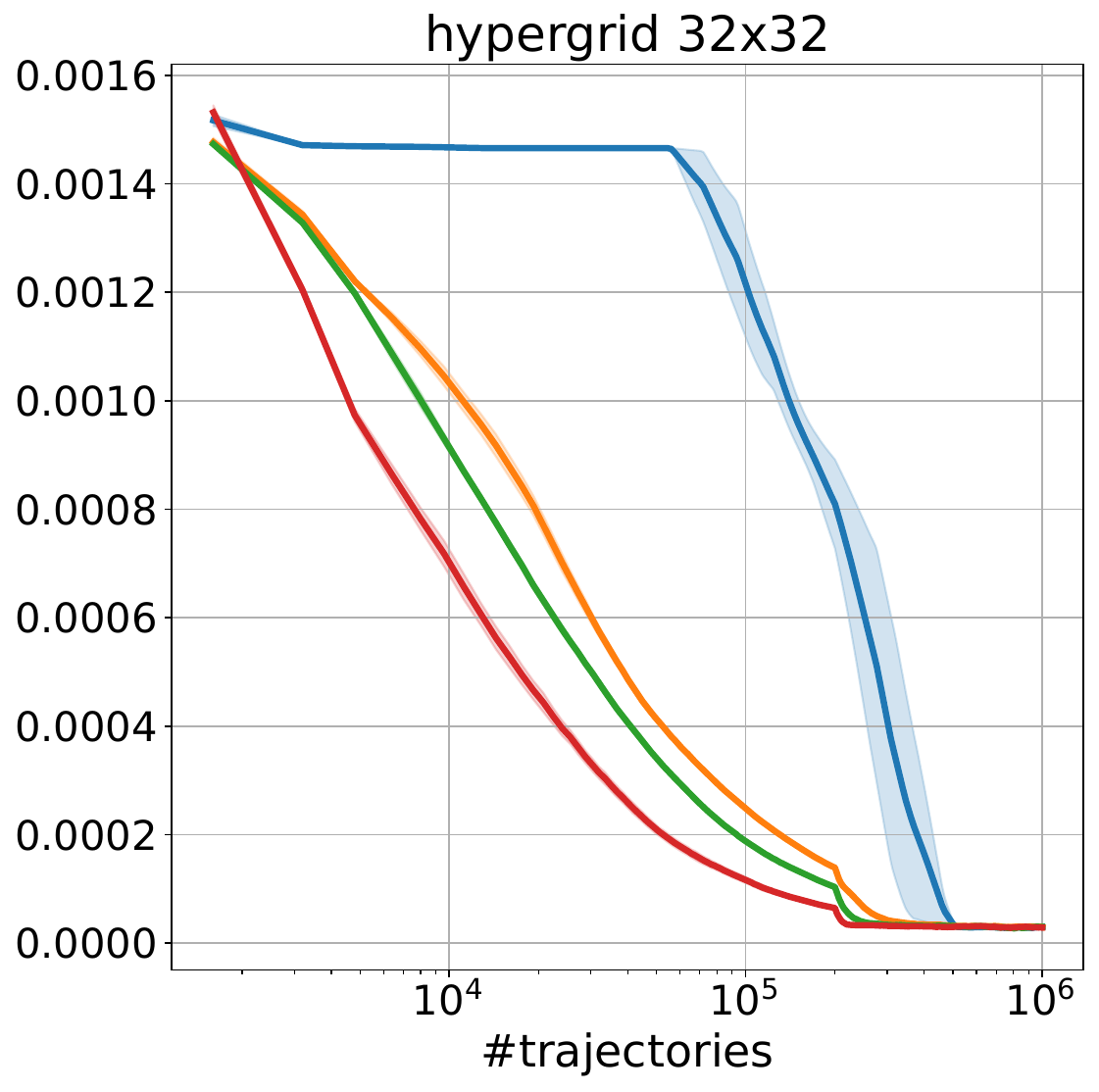}
    \includegraphics[width=0.2334\linewidth]{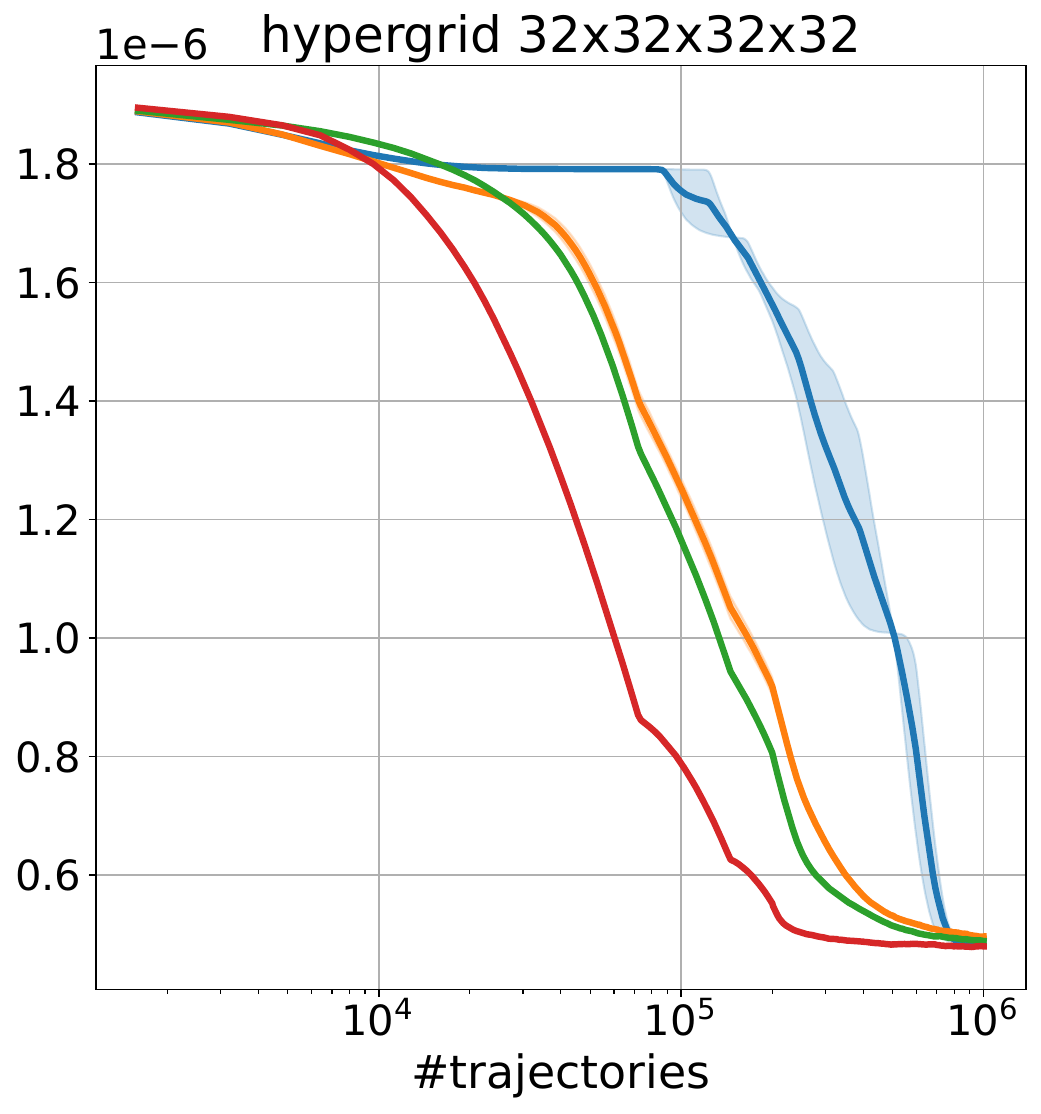}
    \includegraphics[width=0.2599\linewidth]{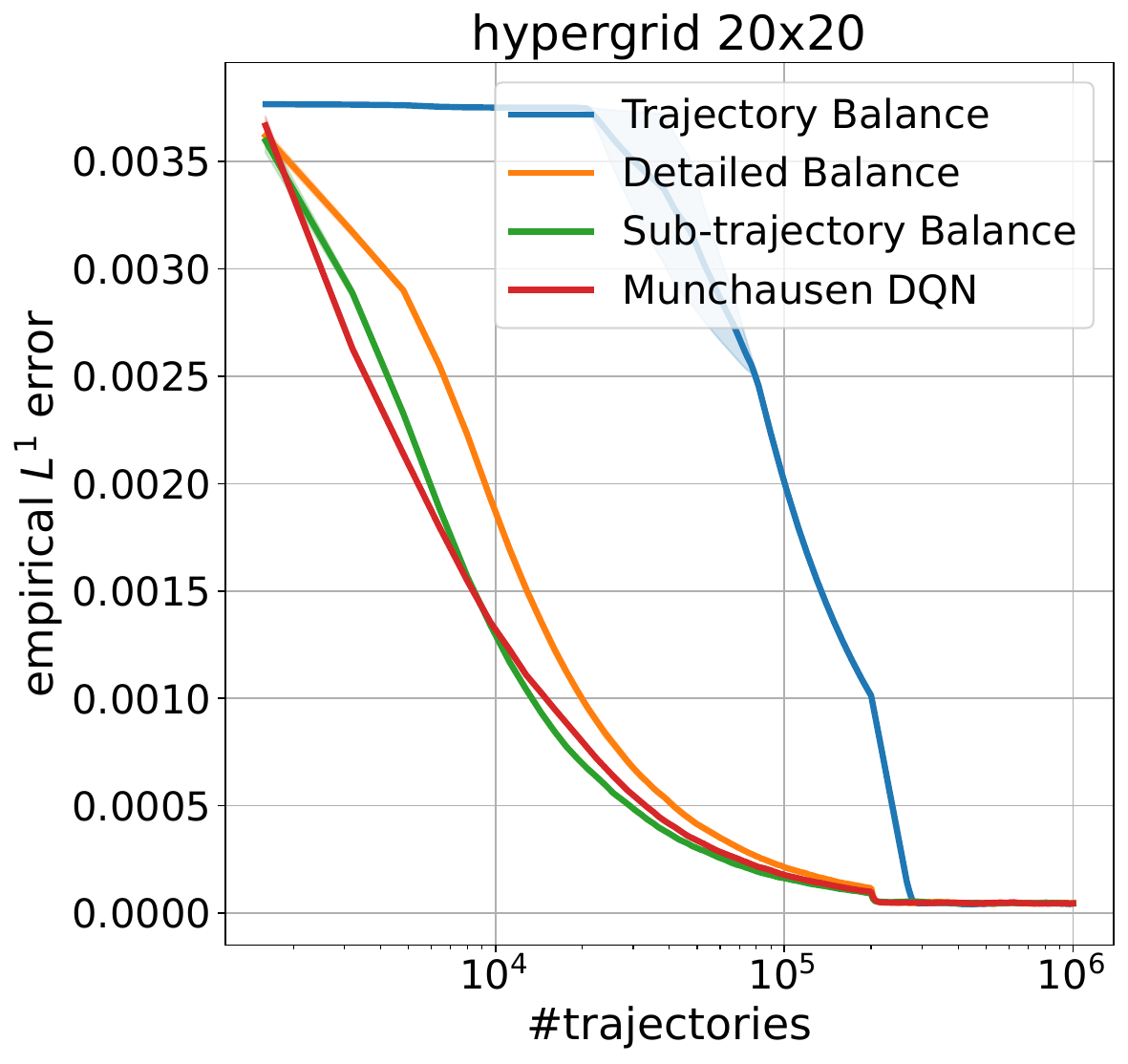}
    \includegraphics[width=0.2290\linewidth]{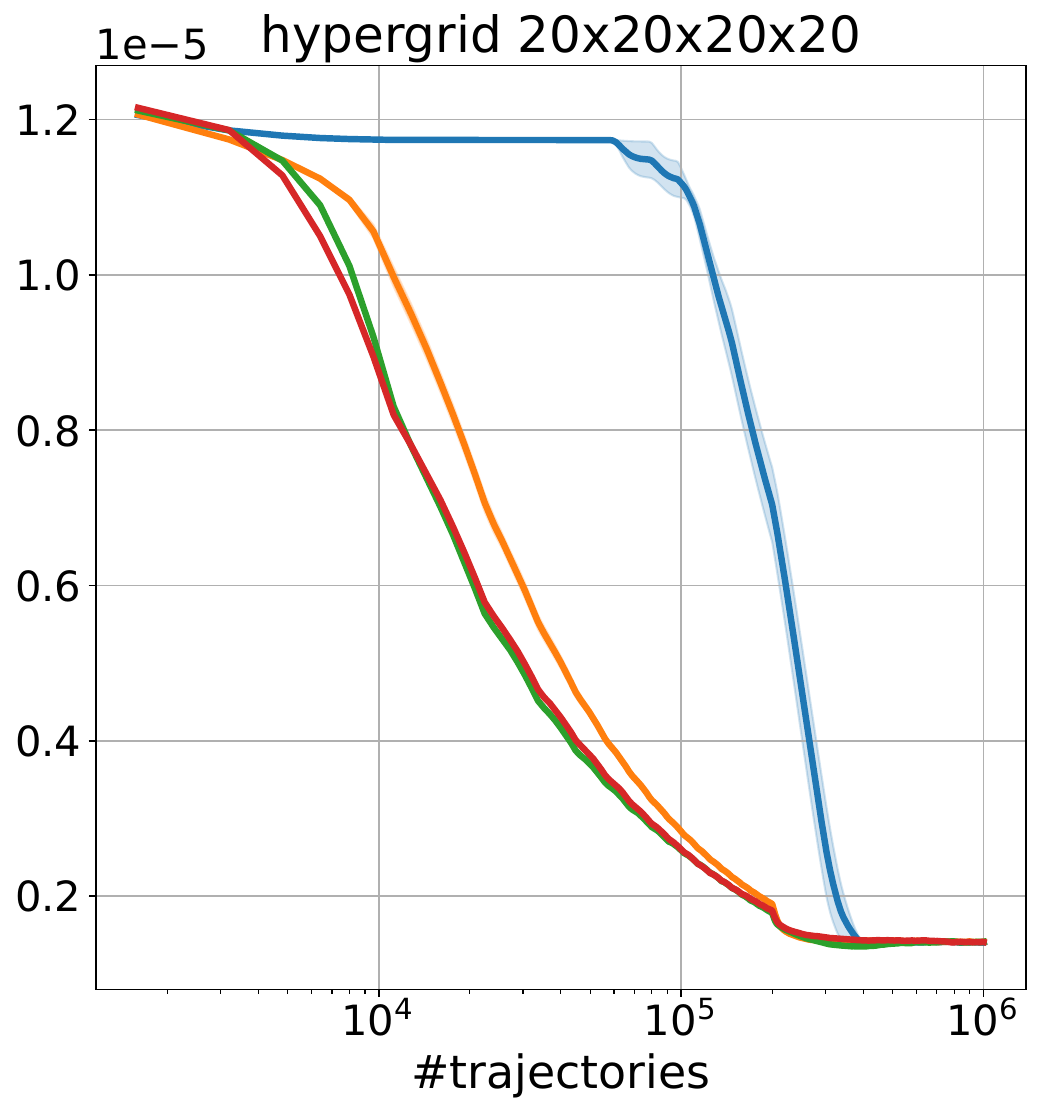}
    \includegraphics[width=0.2508\linewidth]{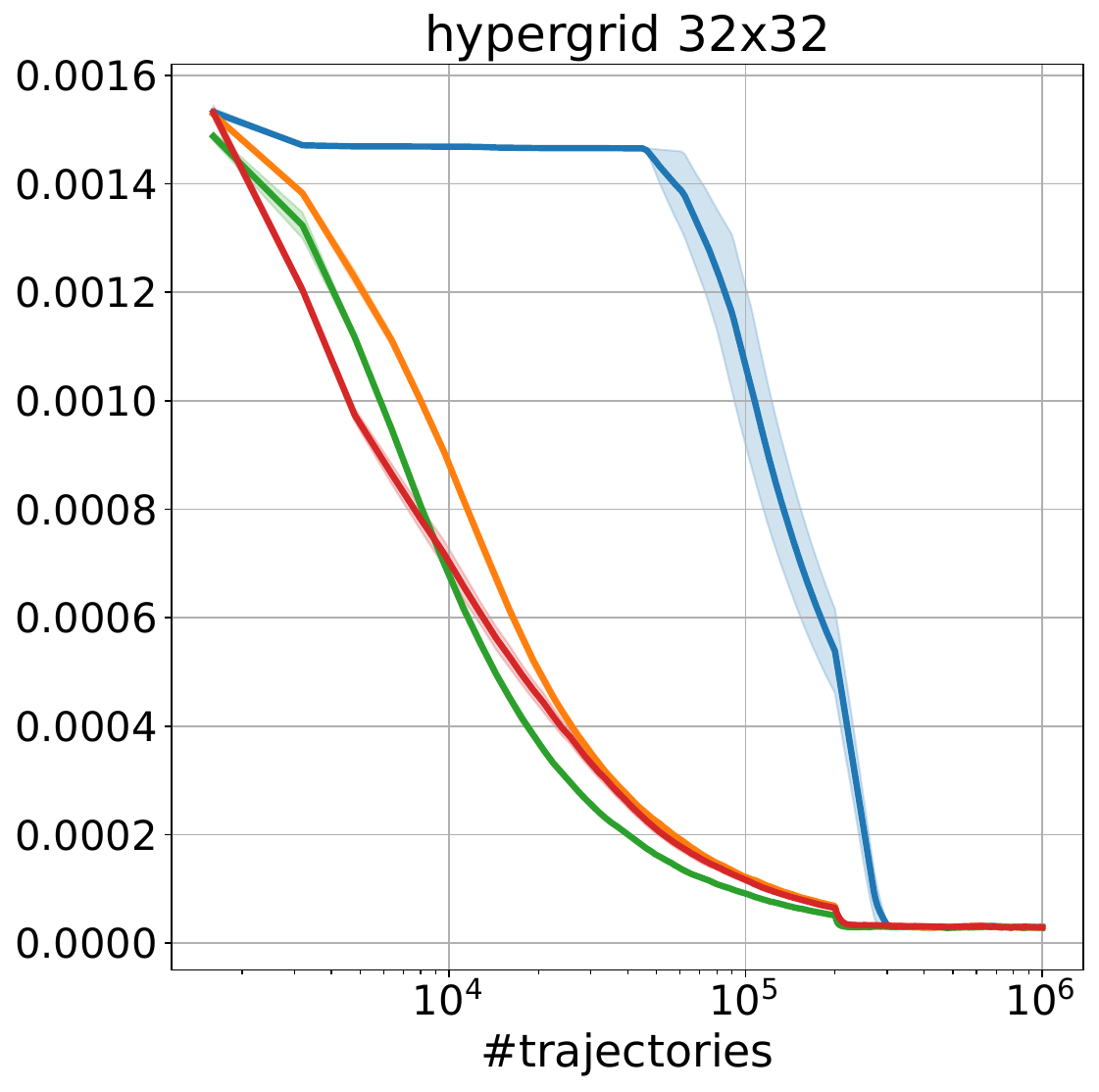}
    \includegraphics[width=0.2334\linewidth]{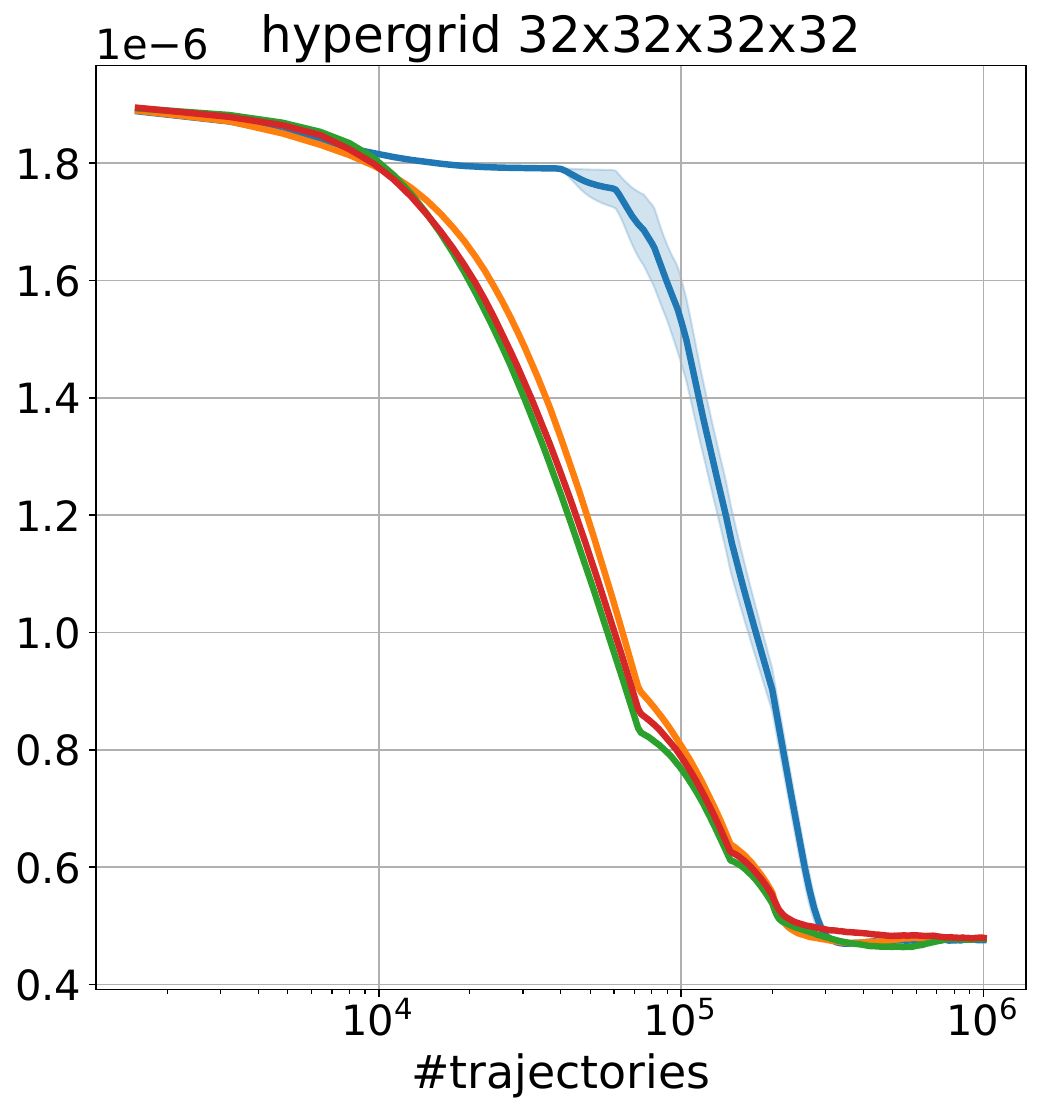}

    \vspace{-0.1cm}
    \caption{$L^1$ distance between target and empirical GFlowNet distributions over the course of training on the hypergrid environment. \textit{Top row:} $\PB$ is fixed to be uniform for all methods. \textit{Bottom row:} $\PB$ is learnt for the baselines and fixed to be uniform for \MDQN. Mean and std values are computed over 3 runs. \vspace{-0.05cm}} 
    \label{fig:hypergrid_base}
    
\end{figure*}

In this section we evaluate the soft RL approach to the task of GFlowNet training under the proposed paradigm and compare it to prior GFlowNet training approaches: trajectory balance (\TB, \citealp{malkin2022trajectory}), detailed balance (\DB, \citealp{bengio2021gflownet}) and subtrajectory balance (\SubTB, \citealp{madan2023learning}). Among soft RL methods, for the sake of brevity this section only presents the results for Munchausen DQN (\MDQN, \citealp{vieillard2020munchausen}) since we found it to show better performance in comparison to other considered techniques. For ablation studies and additional comparisons we refer the reader to Appendix~\ref{app:ablation}.

\vspace{-0.11cm}
\subsection{Hypergrid Environment}\label{sec:grid_experiments}

\vspace{-0.11cm}
We start with synthetic hypergrid environments introduced in~\cite{bengio2021flow}. These environments are small enough that the normalizing constant $\rmZ$ can be computed exactly and the trained policy can be efficiently evaluated against the reward distribution. 

The set of non-terminal states is a set of points with integer coordinates inside a $D$-dimensional hypercube with side length $H$: $\left\{\left(s^1, \ldots, s^D\right) \mid s^i \in\{0,1, \ldots, H-1\}\right\}$.
The initial state is $(0, \ldots, 0)$, and for each non-terminal state $s$ there is its copy $s^{\top}$ which is terminal. For a non-terminal state $s$ the allowed actions are incrementing one coordinate by $1$ without exiting the grid and the terminating action $s \to s^{\top}$. The reward has modes near the corners of the grid which are separated by wide troughs with a very small reward. We study 2-dimensional and 4-dimensional grids with reward parameters taken from~\cite{malkin2022trajectory}. To measure the performance of GFlowNet during training, $L^1$-distance is computed between the true reward distribution $\cR(x) / \rmZ$ and the empirical distribution $\pi(x)$ of last $2 \cdot 10^5$ samples seen in training (endpoints of trajectories sampled from $\PF$): {\normalsize$\frac{1}{|\cX|}\sum_{x \in \cX} |\cR(x) / \rmZ - \pi(x)|$}. Remember, that in case of $\MDQN$ we have $\PF(\cdot \mid s) = \softmax_\lambda(Q_{\theta}(s,\cdot))$ for $\lambda = 1/(1-\alpha)$. All models are parameterized by MLP, which accepts a one-hot encoding of $s$ as input. Additional details can be found in Appendix~\ref{app:exp_grid}.

We consider two cases for the baselines: uniform and learnt $\PB$. Since our theoretical framework does not support learnt $\PB$, we fix it to be uniform for \MDQN in all cases. Figure~\ref{fig:hypergrid_base} presents the results. \MDQN has faster convergence than all baselines in case of uniform $\PB$. Training $\PB$ improves the speed of convergence for the baselines, with \SubTB showing better performance than \MDQN in some cases. However, even with learnt $\PB$, \TB and \DB converge slower than \MDQN.

\vspace{-0.11cm}
\subsection{Small Molecule Generation}\label{sec:mols_experiments}
\vspace{-0.11cm}


Next, we consider the molecule generation task from~\cite{bengio2021flow}. The goal is to generate binders of the sEH (soluble epoxide hydrolase) protein. The molecules represented by graphs are generated by sequentially joining parts from a predefined vocabulary of building blocks~\citep{jin2020junction}. The state space size is approximately $10^{16}$ and each state has between $100$ and $2000$ possible actions. The reward is based on a docking prediction~\citep{trott2010autodock} and is given in a form of pre-trained proxy model $\tilde{\cR}$ from~\cite{bengio2021flow}. Additionally, the reward exponent hyperparameter $\beta$ is introduced, so the GFlowNet is trained using the reward $\cR(x) = \tilde{\cR}(x)^{\beta}$.

We follow the setup from~\cite{madan2023learning}. Each model is trained with different values of learning rate and $\beta$. 
 $\PB$ is fixed to be uniform for all methods. 
To measure how well the trained models match the target distribution, Pearson correlation is computed on a fixed test set of molecules between $\log \cR(x)$ and $\log \cP_{\theta}(x)$, where the latter is log probability that $x$ is the end of a trajectory sampled from $\PF$. $\log \cP_{\theta}(x)$ is estimated the same way as in~\cite{madan2023learning}. Additionally, we track the number of Tanimoto-separated modes captured by each method over the course of training, as proposed in~\cite{bengio2021flow}. Further experimental details are given in Appendix~\ref{app:exp_mol}.

\begin{figure}[t!]

  \centering
  \begin{tabular}{@{}c@{}}
    \includegraphics[width=0.9\linewidth]{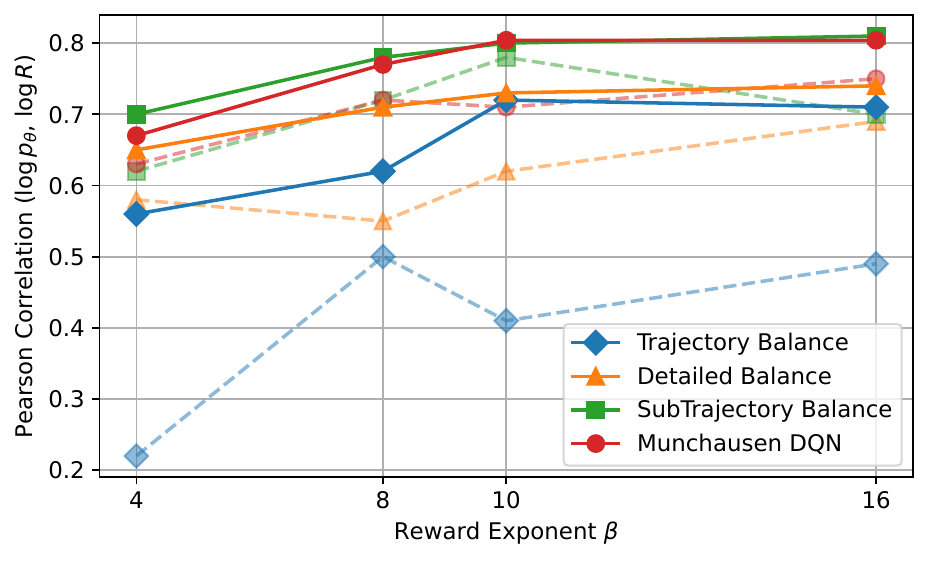} 
  \end{tabular}

  \begin{tabular}{@{}c@{}}
    \includegraphics[width=0.9\linewidth]{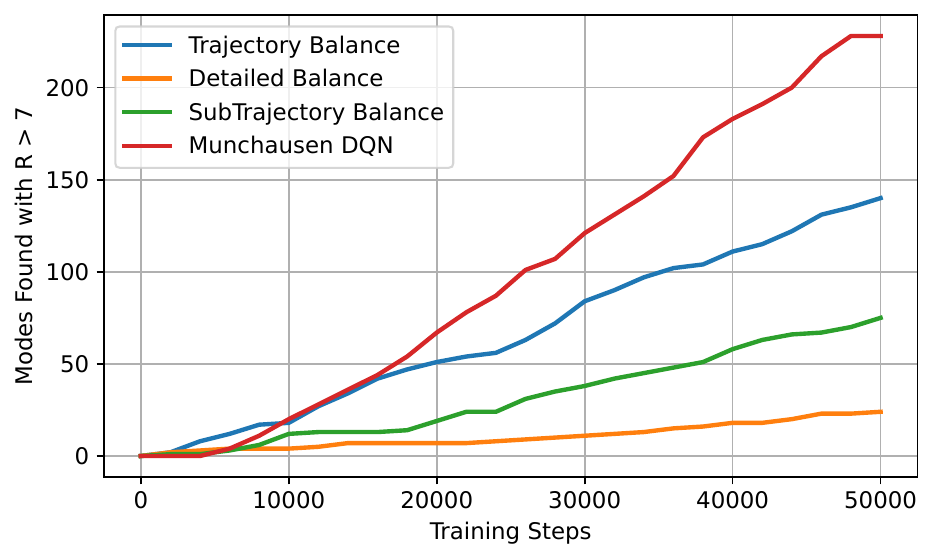} 
  \end{tabular}

    \vspace{-0.15cm}
  \caption{
Small molecule generation results. \textit{Above:} Pearson correlation between $\log \cR$ and $\log \cP_{\theta}$ on a test set for each method and varying $\beta \in \{4, 8, 10, 16\}$. Solid lines represent the best results over choices of learning rate, dashed lines — mean results. \textit{Below:} Number of Tanimoto-separated modes with $\tilde{\cR} > 7.0$ found over the course of training for $\beta = 10$. \vspace{-0.15cm}}\label{fig:mol_results}

\end{figure}

The results are presented in Figure~\ref{fig:mol_results}. We find that \MDQN outperforms \TB and \DB in terms of reward correlation and shows similar (or slightly worse when $\beta=4$) performance when compared to \SubTB. It also has a similar robustness to the choice of learning rate when compared to \SubTB. However, \MDQN discovers more modes over the course of training than all baselines.

\vspace{-0.15cm}
\subsection{Non-Autoregressive Sequence Generation}\label{sec:bit_experiments}

\vspace{-0.15cm}
Finally, we consider the bit sequence generation task from~\cite{malkin2022trajectory} and~\cite{madan2023learning}. To create a harder GFlowNet learning task with a non-tree DAG structure, we modify the state and action spaces, allowing non-autoregressive generation akin to the approach used in~\cite{zhang2022generative}.

\begin{figure}[t!]

  \centering
  \begin{tabular}{@{}c@{}}
    \includegraphics[width=0.9\linewidth]{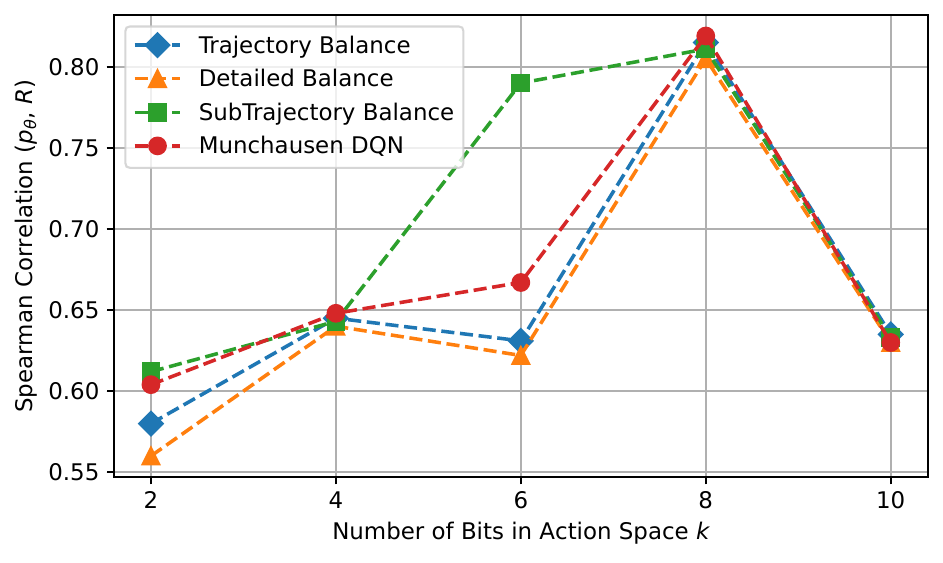} 
  \end{tabular}

  \begin{tabular}{@{}c@{}}
    \includegraphics[width=0.9\linewidth]{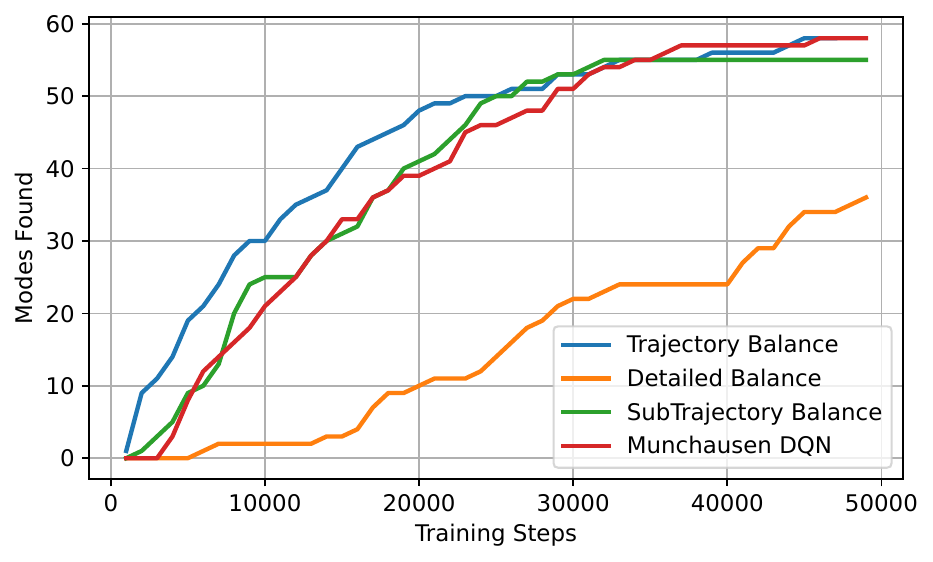} 
  \end{tabular}

    \vspace{-0.15cm}
  \caption{Bit sequence generation results. \textit{Above:} Spearman correlation between $\cR$ and $\cP_{\theta}$ on a test set for each method and varying $k \in \{2, 4, 6, 8, 10\}$. \textit{Below:} The number of modes discovered over the course of training for $k = 8$. \vspace{-0.17cm}}\label{fig:bit_results}

\end{figure}

The goal is to generate binary strings of a fixed length $n = 120$. The reward is specified by the set of modes $M \subset \cX = \{0, 1\}^n$ and is defined as $\cR(x) = \exp(-\min_{x' \in M} d(x, x'))$, where $d$ is Hamming distance. We construct $M$ the same way as in~\cite{malkin2022trajectory} and use $|M| = 60$. Parameter $k$ is introduced as a tradeoff between trajectory length and action space sizes. For varying $k \mid n$, the vocabulary is defined as a set of all $k$-bit words. The generation starts at a sequence of $n/k$ special "empty" words $\oslash$, and at each step the allowed actions are to pick any position with an empty word and replace it with some word from the vocabulary. Thus the state space consists of sequences of $n/k$ words, having one of $2^k + 1$ possible words at each position. The terminal states are the states with no empty words, thus corresponding to binary strings of length $n$. 

To evaluate the model performance, we utilize two metrics used in~\cite{malkin2022trajectory} and~\cite{madan2023learning} for this task: 1) test set Spearman correlation between $\cR(x)$ and $\cP_{\theta}(x)$ 2) number of modes captured during training (number of bit strings
from $M$ for which a candidate within a distance $\delta = 30$ has been generated). In contrast to the setup in~\cite{malkin2022trajectory}, the exact probability of sampling $x$ from GFlowNet is intractable in our case due to a large number of paths leading to each $x$, so we utilize a Monte Carlo estimate proposed in~\cite{zhang2022generative}. We fix $\PB$ to be uniform in our experiments. Additional details can be found in Appendix~\ref{app:exp_bit}. 

Figure~\ref{fig:bit_results} presents the results. \MDQN has similar or better reward correlation than all baselines except in the case of $k = 6$, where it outperforms \TB and \DB but falls behind \SubTB. In terms of discovered modes, it shows better performance than \DB and similar performance to \TB and \SubTB.

\vspace{-0.15cm}
\section{CONCLUSION}\label{sec:conclusion}

\vspace{-0.15cm}
In this paper we established a direct connection between GFlowNets and entropy-regularized RL in the general case and experimentally verified it. If a GFlowNet backward policy is fixed, an MDP is appropriately constructed from a GFlowNet DAG and specific rewards are set for all transitions, we showed that learning the GFlowNet forward policy $\PF$ is equivalent to learning the optimal RL policy $\pistar_\lambda$ when entropy regularization coefficient $\lambda$ is set to $1$. In our experiments, we demonstrated that \MDQN \citep{vieillard2020munchausen} shows competitive results when compared to established GFlowNet training methods~\citep{malkin2022trajectory, bengio2021gflownet, madan2023learning} and even outperforms them in some cases.

A promising direction for further research is to explore theoretical properties of GFlowNets by extending the analysis of  \cite{wen2017efficient}. Additionally, it would be valuable to generalize the established connection beyond the discrete setting, following the work of \cite{lahlou2023theory}. Furthermore, from a practical perspective there is an interesting direction of extending Monte-Carlo Tree Search-based (MCTS) algorithms to GFlowNets such as AlphaZero \citep{silver2017general} since known deterministic environments allow to perform MCTS rollouts. Finally, in the context of existing connections between GFlowNets and 1) generative modeling \citep{zhang2022unifying}, 2) Markov Chain Monte Carlo methods \citep{deleu2023generative}, 3) variational inference~\citep{zimmermann2022variational, malkin2023gflownets}, with the addition of the presented link to 4) reinforcement learning, another potential direction can be to view GFlowNets as a unifying point of different ML areas and models, which can lead to further establishment of valuable connections.

\section*{Acknowledgements}


The experimental results from Section~\ref{sec:experiments} were obtained by Nikita Morozov and Alexey Naumov with the support of the grant for research centers in the field of AI provided by the Analytical Center for the Government of the Russian Federation (ACRF) in accordance with the agreement on the provision of subsidies (identifier of the agreement 000000D730321P5Q0002) and the agreement with HSE University No. 70-2021-00139. Theoretical results from Section~\ref{sec:representation} were obtained by Daniil Tiapkin with the support by the Paris Île-de-France Région in the framework of DIM AI4IDF. This research was supported in part through computational resources of HPC facilities at HSE University~\citep{kostenetskiy2021hpc}.

\bibliographystyle{apalike}
\bibliography{ref}

\clearpage
\section*{Checklist}

 \begin{enumerate}

  \item For all models and algorithms presented, check if you include:
  \begin{enumerate}
    \item A clear description of the mathematical setting, assumptions, algorithm, and/or model. [Yes, see Section~\ref{sec:background} and \ref{sec:representation}, also Appendix~\ref{app:algo}]
    \item An analysis of the properties and complexity (time, space, sample size) of any algorithm. [Yes, see Appendix~\ref{app:algo}]
    \item (Optional) Anonymized source code, with specification of all dependencies, including external libraries. [Yes, see Section~\ref{sec:intro}]
  \end{enumerate}

  \item For any theoretical claim, check if you include:
  \begin{enumerate}
    \item Statements of the full set of assumptions of all theoretical results. [Yes, see Section~\ref{sec:representation}]
    \item Complete proofs of all theoretical results. [Yes, see Appendix~\ref{app:theory}]
    \item Clear explanations of any assumptions. [Yes, see Section~\ref{sec:representation}]     
  \end{enumerate}

  \item For all figures and tables that present empirical results, check if you include:
  \begin{enumerate}
    \item The code, data, and instructions needed to reproduce the main experimental results (either in the supplemental material or as a URL). [Yes, see Section~\ref{sec:intro} and Appendix~\ref{app:exp}]
    \item All the training details (e.g., data splits, hyperparameters, how they were chosen). [Yes, see Section~\ref{sec:experiments} and Appendix~\ref{app:exp}]
          \item A clear definition of the specific measure or statistics and error bars (e.g., with respect to the random seed after running experiments multiple times). [Yes, see Section~\ref{sec:experiments} and Appendix~\ref{app:exp}]
          \item A description of the computing infrastructure used. (e.g., type of GPUs, internal cluster, or cloud provider). [Yes, see Appendix~\ref{app:exp}]
  \end{enumerate}

  \item If you are using existing assets (e.g., code, data, models) or curating/releasing new assets, check if you include:
  \begin{enumerate}
    \item Citations of the creator If your work uses existing assets. [Yes, see Appendix~\ref{app:exp}]
    \item The license information of the assets, if applicable. [Not Applicable]
    \item New assets either in the supplemental material or as a URL, if applicable. [Not Applicable]
    \item Information about consent from data providers/curators. [Not Applicable]
    \item Discussion of sensible content if applicable, e.g., personally identifiable information or offensive content. [Not Applicable]
  \end{enumerate}

  \item If you used crowdsourcing or conducted research with human subjects, check if you include:
  \begin{enumerate}
    \item The full text of instructions given to participants and screenshots. [Not Applicable]
    \item Descriptions of potential participant risks, with links to Institutional Review Board (IRB) approvals if applicable. [Not Applicable]
    \item The estimated hourly wage paid to participants and the total amount spent on participant compensation. [Not Applicable]
  \end{enumerate}

 \end{enumerate}

\newpage
\onecolumn
\appendix

\section{NOTATION}\label{app:notation}

\begin{table}[h]
	\centering
	\caption{Table of notation used throughout the paper.}
	\begin{tabular}{@{}l|l@{}}
		\toprule
		\thead{Notation} & \thead{Meaning} \\ \midrule
	$\cS$ & state space\\
	$\cA$ & action space\\
	$\cX$ & sampling space\\
	$\cR$ & GFlowNet reward function \\
	$\cF(\tau), \cF(s), \cF(s\to s')$ & flow function \\
    $\cP(\tau)$ & distribution over trajectories induced by a flow \\
    $\cP(x)$ & distribution over terminal states induced by a flow \\
    $\PF(s'|s)$ & GFlowNet forward policy \\
    $\PB(s|s')$ & GFlowNet backward policy \\
    $\rmZ, \rmZ_{\cF}$ & normalizing constant of the reward distribution and a flow $\cF$ \\
	\hline
	$\MK(s'|s,a)$ & Markovian transition kernel for MDP \\
	$r(s,a)$ & MDP reward function \\
    $\lambda$ & regularization coefficient \\
    $V^{\pi}_{\lambda}(s), Q^{\pi}_{\lambda}(s,a)$ & regularized value and Q-value of a given policy $\pi$\\
	$\Vstar_{\lambda}(s), \Qstar_{\lambda}(s,a)$ & regularized value and Q-value of an optimal policy \\
    $V^{\pi}_1(s), Q^{\pi}_1(s,a)$ & regularized value and Q-value of a given policy $\pi$ with coefficient $1$\\
	$\Vstar_1(s), \Qstar_1(s,a)$ & regularized value and Q-value of an optimal policy  with coefficient $1$ \\
    $\pistar_{\lambda}(a|s)$ & regularized optimal policy \\
    $q^\pi(\tau)$ & distribution over trajecories induced by a policy $\pi $\\
    $d^\pi(x)$ & distribution over terminal states induced by a policy $\pi$ \\
    \hline 
    $\softmax_{\lambda}$ & softmax function with a temperature $\lambda > 0$, for $x\in \R^d$ \\
    & $[\softmax_{\lambda}(x)]_i \triangleq \rme^{x_i /\lambda}/(\sum_{j=1}^d \rme^{x_j / \lambda} )\ \forall i \in [d]$ \\
    $\logsumexp_{\lambda}$ & log-sum-exp function with a temperature $\lambda > 0$, for $x \in \R^d$ \\
    &$\logsumexp(x) \triangleq \lambda \log(\sum_{i=1}^d \rme^{x_i / \lambda})$. \\
    $\cH$ & Shannon entropy function, $\cH(p) \triangleq \sum_{i=1}^n p_i \log(1/p_i)$.  \\
     \bottomrule
	\end{tabular}
\end{table}

Let $(\Xset,\Xsigma)$ be a measurable space and $\Pens(\Xset)$ be the set of all probability measures on this space. For $p \in \Pens(\Xset)$ we denote by $\E_p$ the expectation w.r.t.\,$p$. For random variable $\xi: \Xset \to \R$ notation $\xi \sim p$ means $\operatorname{Law}(\xi) = p$. 

For any $p, q \in \Pens(\Xset)$ the Kullback-Leibler divergence $\KL(p, q)$ is given by
$$
\KL(p, q) \triangleq \begin{cases}
\E_{p}\left[\log \frac{\rmd p}{\rmd q}\right], & p \ll q, \\
+ \infty, & \text{otherwise.}
\end{cases} 
$$


\section{MISSING PROOFS}\label{app:theory}

\subsection{Proof of Theorem~\ref{th:gflownet_reduction}}
    Let $\cF \colon \cT \to \R_{\geq 0}$ be a Markovian flow uniquely defined by a backward policy $\PB$ and a GFlowNet reward $\cR$ that satisfies reward matching constraint \eqref{eq:reward_matching}. By soft Bellman equations \eqref{eq:soft_bellman_equations}, the value and the Q-value of the optimal policy satisfy for all $s \in \cS \setminus \cX $,$s' \in \cA_s$
    \begin{align}
        \begin{split}\label{eq:gfn_bellmean_equations}
            \Qstar_1(s,s') &= r(s,s') + \Vstar_1(s'), \\
         \Vstar_1(s) &= \log\left( \sum_{s' \in \cA_s} \exp(\Qstar_1(s,s'))\right)  
        \end{split}
    \end{align}
    and, additionally, $\Vstar_1(s_f) = 0$ and $\Vstar_1(x) = \log \cR(x)$ for any $x \in \cX$. 
    
    Next, we claim that for any $s,s' \in \cS$ it holds $\Vstar_1(s) = \log \cF(s)$, $\Qstar_1(s,s') = \log \cF(s\to s')$ and prove it by backward induction over the topological orderering of vertices of $\cG$. The base of induction already holds for $\Vstar_1(x)$ for all $x \in \cX$. Let $s \in \cS \setminus \cX$ be a fixed state and assume that the claim is proven for all its children $s' \in \cA_s$. By soft Bellman equations \eqref{eq:gfn_bellmean_equations}, we have for any $s' \in \cA_s$

    \begin{align*}
        \Qstar_1(s,s') &= \log \PB(s| s') + \Vstar_1(s') \\
        &= \log \frac{\cF(s \to s')}{\cF(s')} + \log \cF(s') = \log \cF(s\to s'),
    \end{align*}
    where we used the choice of MDP rewards \eqref{eq:def_mdp_reward} and the definition of $\PB$ through the flow function \eqref{eq:pf_pb_definition} combined with the induction hypothesis. Next, soft Bellman equations imply that

    \begin{align*}
        \exp(\Vstar_1(s)) &= \sum_{s' \in \cA_s} \exp({\Qstar_1(s,s')}) = \sum_{s' \in \cA_s} \cF(s\to s') = \cF (s),
    \end{align*}
    where the induction hypothesis for Q-values and flow machining constraint \citep[Proposition 19]{bengio2021gflownet} are applied. Finally, using \eqref{eq:soft_optimal_policy} we have the following expression for the optimal policy
    \[
        \pistar_1(s'|s) = \exp\left( \Qstar_1(s,s') - \Vstar_1(s) \right) = \frac{\cF(s \to s')}{\cF(s)}
    \]
    that coincides with the forward policy $\PF$ \eqref{eq:pf_pb_definition} corresponding to the flow.

    
\subsection{Proof of Proposition~\ref{prop:value_expression}}
    Let us take $N$ equal to the maximal length of trajectory $\max_{\tau} n_{\tau}$, i.e. at step $N$ agent is guaranteed to reach the absorbing state with zero reward. Then, by the definition \eqref{eq:regularized_value_def} of $V^{\pi}_1(s_0)$ in a regularized MDP with  $\gamma = \lambda=1$ we have
    \[
        V^{\pi}_1(s_0) = \E_{\pi}\left[ \sum_{t=0}^{N-1} r(s_t,s_{t+1}) - \log \pi(s_{t+1}| s_t) \right],
    \]
    where we used the tower property of conditional expectation to replace entropy with negative logarithm. By substituting the expression for rewards we have
    \begin{align*}
        V^{\pi}_1(s_0) &= \E_{\tau \sim q^{\pi}}\left[ \sum_{t=1}^{n_{\tau}} \log \frac{\PB(s_{t-1} | s_{t})}{\pi(s_{t} | s_{t-1})} + \log \cR(s_{n_{\tau}})  \right]\\
        &= \E_{\tau \sim q^{\pi}}\left[ \log \frac{\prod_{t=1}^{n_{\tau}} \PB(s_{t-1} |s_t)  \cdot \cR(s_{n_{\tau}}) \cdot \rmZ }{\prod_{t=1}^{n_{\tau}} \pi(s_t | s_{t-1}) \cdot \rmZ } \right] = - \KL(q^{\pi} \Vert \PB) + \log \rmZ.
    \end{align*}



\section{DETAILED ALGORITHM DESCRIPTION}\label{app:algo}

In this section we provide a detailed algorithmic description of \SoftDQN and \MunDQN and discuss the effects of several hyperparameters.

\subsection{Soft DQN}

The pipeline of this algorithm is as follows. In the beginning of the learning procedure we have two neural networks: $Q_\theta$ and $Q_{\bar \theta}$, and $\bar \theta = \theta$.

\begin{algorithm}[!t]
\centering
\caption{\SoftDQN \citep{haarnoja2017reinforcement} / \textcolor{blue}{\MunDQN \citep{vieillard2020munchausen}}}
\label{alg:softDQN}
\begin{algorithmic}[1]
  \STATE {\bfseries Input:} trajectory budget $N$, number of trajectories on one update $M$, exploration parameter $\varepsilon$, backward kernel $\PB$, PER $\cB$, PER batch size $B$, hard update parameter $T$, soft update parameter $\tau$, \textcolor{blue}{Munchausen parameters $\alpha$ ,$l_0$};
  \STATE Initialize parameters of online network $\theta$ and target network $\bar \theta = \theta$;
  \STATE Define entropy coefficient $\lambda = 1\  (\textcolor{blue}{1/(1-\alpha)})$
      \FOR{ $t = 1,\ldots, \lceil N/M \rceil$}
        \STATE Sample $M$ trajectories interaction with $\varepsilon$-greedy version of policy $\pi_\theta = \softmax_{\lambda}(Q_\theta)$, and place them into replay buffer $\cB$;
        \STATE Sample $B$ transitions $\{ (s_k, a_k, R_k, s'_k, d_k) \}_{k=1}^N$ from PER $\cB$, where $d_k$ is a \texttt{DONE} flag: $d_k = \ind\{ s_k \in \cX \}$, and $R_k$ is GflowNet reward is $d_k=1$ and $0$ otherwise;
        \STATE Compute MDP rewards: $r_k = d_k \log R_k + (1-d_k) \log \PB(s_k | s'_k)$ with a convention $0\log 0 = 0$ for all $k \in \{1,\ldots,B\}$.
        \STATE Compute TD targets
        \[
            y_k = r_k + (1- d_k) \logsumexp_{\lambda}\left( Q_{\bar \theta}(s'_k) \right) \textcolor{blue}{+  \alpha  \max\{ \lambda \log \pi_{\bar \theta}(a_k | s_k), l_0\}}
        \]
        optionally using a convention $\logsumexp_{\lambda}\left( Q_{\bar \theta}(s'_k) \right) = \log \cR(s'_k)$ $\forall s'_k \in \cX$ (see explanation below);
        \STATE Use $y_k$ to update priorities in PER \citep{schaul2016prioritized} as
        \[
            p_k =  \ell\left( Q_{\theta}(s_k, a_k), y_k \right),
        \]
        where $\ell $ is a Huber loss \eqref{eq:huber_loss}.
       \STATE Compute TD loss
       \[
            \cL(\theta) = \frac{1}{B} \sum_{k=1}^B \ell\left( Q_{\theta}(s_k, a_k), y_k \right),
       \]
       where $\ell$ is a Huber loss \eqref{eq:huber_loss}.
       \STATE Perform gradient update by $\nabla_\theta \cL(\theta)$;
        \IF{$t \mod T = 0$}
            \STATE Update target network $\bar\theta := (1-\tau) \bar \theta + \tau \theta$;
        \ENDIF
    \ENDFOR
    \STATE \textbf{Output} policy $\pi_\theta$.
\end{algorithmic}
\end{algorithm}

On each iteration of the algorithm, several trajectories (usually 4 or 16) are sampled by following a softmax policy $\pi_\theta(s) = \softmax_{\lambda}(Q_{\theta}(s))$ (here $\lambda=1$) using an online network $Q_\theta$ with (optionally) an additional $\varepsilon$-greedy exploration.\footnote{Notably, softmax parameterization naturally induces Boltzman exploration. However, the effects of Boltzman and $\varepsilon$-exploration are different and in general cannot substitute each other, see discussion in \citep{vieillard2020munchausen}.}

All these sampled trajectories are separated onto transitions $(s_t, a_t, s_{t+1}, R_t, d_t)$, where $R_t = \log \cR(s_t)$ if $s_t \in \cX$ and zero otherwise, $d_t = \ind\{ s_t \in \cX \}$ is a \texttt{DONE} float. Note that we always have $a_t = s_{t + 1}$, and $a_t$ is used solely for the purpose of consistency with common RL notation. These separated transitions are placed into a replay buffer $\cB$ \citep{lin1992self} with a prioritization mechanism following \cite{schaul2016prioritized}.

Next, a batch of transitions $\{ (s_k, a_k, s'_k, R_k, d_k) \}_{k=1}^B$ is sampled from the replay buffer $\cB$ using prioritization. MDP rewards are computed as $r_k = d_k \log \cR(s_k) + (1 - d_k) \log \PB(s_k | s'_k)$, and TD-targets are computed as follows:
\begin{equation}\label{eq:td_target_softdqn}
    \forall k \in [B]: y_k = r_k + (1-d_k) \logsumexp_{\lambda}(Q_{\bar \theta}(s'_k)),
\end{equation}
where $\lambda = 1$. 

Alternatively, one can fully skip transitions that lead to the sink state $s_f$ (transitions for which $d_k = 1$), and instead substitute $\logsumexp_{\lambda}(Q_{\bar \theta}(s'_k)) = \log \cR(s'_k)$ $\forall s'_k \in \cX$. This can be done since we know the exact value of $\Vstar_{\lambda}(s)$ for all $s \in \cX$. We follow this choice in our experiments.


Given targets, we update priorities for a sampled batch in the replay buffer and compute the regression TD-loss as follows:
\[
    \cL(\theta) = \frac{1}{B} \sum_{k=1}^B \ell(Q_{\theta}(s_k,a_k), y_k),
\]
where $\ell$ is a Huber loss
\begin{equation}\label{eq:huber_loss}
    \ell(x,y) = \begin{cases}
        \frac{1}{2} (x-y)^2 & |x-y| \leq 1 \\
        |x-y| - 1/2 & |x-y| \geq 1.
    \end{cases}
\end{equation}

Finally, at the end of each iteration, we perform either a hard update by copying weights of an online network to weights of the target network following the original \DQN approach \citep{mnih2015human} or a soft update by applying Polyak averaging with a parameter $\tau < 1$ following \texttt{DDPG} approach \citep{silver2014deterministic,lillicrap2015continuous}.

The detailed description of this algorithm is also provided in Algorithm~\ref{alg:softDQN}. Notice that the complexity of this algorithm is linear in the number of sampled transitions since the size of the buffer is fixed during the training.

Next, we discuss some elements of this pipeline in more detail.

\paragraph{Prioritized experience replay}
The essence of prioritized experience replay by \citep{schaul2016prioritized} is to sample observations from a replay buffer not uniformly but according to priorities that are computed as follows:
\[
    \mathrm{P}(i) = \frac{|p_i|^\alpha}{\sum_{i \in \cB} |p_i|^{\alpha}},
\]
where $\mathrm{P}(i)$ is a probability to sample $i$-th transition in the replay buffer $\cB$, $p_i$ is a priority that is equal to the last computed TD-error for this transition, and $\alpha$ is a temperature parameter that interpolates between uniform sampling $\alpha = 0$ and pure prioritization $\alpha = +\infty$. A typically chosen value of $\alpha$ is between $0.4$ and $0.6$, however, for two experiments (molecule and bit sequence generation) we found it more beneficial to select an aggressive value $\alpha = 0.9$ to collect hard examples faster.

Additionally, authors of \citep{schaul2016prioritized} applied importance-sampling (IS) correction since the theory prescribes to sample uniformly over the buffer due to stochastic next state $s'_k$. Smoothed correction is controlled with a coefficient $\beta$ that should be annealed to $1$ over the course of training. 

However, in the setup of GFlowNets we noticed that this correction is not needed since the next transition and reward are always deterministic; therefore it is possible to set $\beta$ to be a small constant without any loss of convergence properties. We set $\beta = 0.0$ in the hypergrid experiment and $\beta = 0.1$ in the molecule and bit sequence generation experiments and found it more beneficial than doing higher IS-correction.

\paragraph{Choice of loss}
We examine two types of regression losses: classic MSE loss and Huber loss that was applied in \DQN \citep{mnih2015human} and \MunDQN \citep{vieillard2020munchausen}.  The main idea behind the Huber loss is automatic gradient clipping for very large updates. We observed such benefits as more stable training while using Huber loss, so we utilized it in all our experiments except some ablation studies in Appendix~\ref{app:ablation}.

\subsection{Munchausen DQN}

The differences between \SoftDQN and \MDQN in the context of GFlowNets are:
\begin{itemize}
    \item Two additional parameters: $\alpha$ and truncation parameter $l_0$;
    \item Using entropy coefficient $\lambda$ equal to $1/(1-\alpha)$ instead of just $\lambda = 1$. In particular, it changes how policy $\pi_{\theta}(s) = \softmax_{1/(1-\alpha)}(Q_\theta(s))$ and value $V_{\theta}(s) = \logsumexp_{1/(1-\alpha)} (Q_\theta(s))$ are computed;
    \item Additional term $\alpha  \max\{ \lambda \log \pi_{\bar \theta}(a_k | s_k), l_0\}$ in the computation of TD-target \eqref{eq:td_target_softdqn}.
\end{itemize}

All these differences are highlighted in Algorithm~\ref{alg:softDQN} with a \textcolor{blue}{blue} color. The use of a different entropy coefficient is justified by Theorem~1 of \cite{vieillard2020munchausen}: \MDQN with entropy coefficient $\lambda$ and Munchausen coefficient $\alpha$ is equivalent to solving entropy-regularized RL with coefficient $(1-\alpha)\lambda$. 

\section{EXPERIMENTAL DETAILS}\label{app:exp}

In this section we provide additional details for the experiments from Section~\ref{sec:experiments}.

\subsection{Hypergrids}\label{app:exp_grid}

The reward at $s^{\top} = (s^1, \ldots, s^D)^{\top}$ is formally defined as
\begin{align*}
\cR(s^{\top}) = R_0 &+ R_1 \prod_{i = 1}^D \mathbb{I}\left[0.25 < \left|\frac{s^i}{H-1}-0.5\right|\right] \\ 
&+ R_2 \prod_{i = 1}^D \mathbb{I}\left[0.3 < \left|\frac{s^i}{H-1}-0.5\right| < 0.4\right],
\end{align*}
where $0<R_0 \ll R_1<R_2$. This reward has modes near the corners of the grid which are separated by wide troughs with a very small reward of $R_0$. In Section~\ref{sec:experiments} we use reward parameters $(R_0 = 10^{-3}, R_1 = 0.5, R_2 = 2.0)$ taken from~\cite{malkin2022trajectory}.

All models are parameterized by MLP of the same architecture as in~\cite{bengio2021flow} with 2 hidden layers and 256 hidden size. Following~\cite{malkin2022trajectory}, we train all models with Adam optimizer and use a learning rate of $10^{-3}$ and a batch size 16 (number of trajectories sampled at each training step). In case of \TB, we use a learning rate of $10^{-1}$ for $\rmZ_{\theta}$ following~\cite{malkin2022trajectory}. For \SubTB parameter $\lambda$ (not to be confused with soft RL regularization coefficient) we use the value of $0.9$ following~\cite{madan2023learning}. All models are trained until $10^6$ trajectories are sampled and the empirical sample distribution $\pi(x)$ is computed over the last $2 \cdot 10^{5}$ samples seen in training, which explains the drop in plots at the $2 \cdot 10^{5}$ mark (see Figure~\ref{fig:hypergrid_base}). 

For \MDQN we use Huber loss instead of MSE, following \cite{mnih2015human,vieillard2020munchausen} and prioritized replay buffer \citep{schaul2016prioritized} from \texttt{TorchRL} library \citep{bou2023torchrl} of size $100,000$, a standard constant $\alpha=0.5$ and without IS correction $\beta =0$. At each training step, we sample 16 trajectories, put all transitions to the buffer, and then sample 256 transitions from the buffer to calculate the loss. Thus \MDQN uses the same number of trajectories and the same number of gradient steps for training as the baselines. Regarding specific Munchausen parameters \citep{vieillard2020munchausen}, we use $\alpha=0.15$ and unusually large $l_0 = -100$, since the entropy coefficient for our setup is much larger than one that was used in the original \MDQN. To update the target network, we utilize soft updates akin to celebrated \texttt{DDPG} \citep{silver2014deterministic,lillicrap2015continuous} with a large coefficient $\tau = 0.25$ due to sampling $16$ trajectories prior to obtaining a batch. 

For hypergrid environment experiments we utilized \texttt{torchgfn} library~\citep{lahlou2023torchgfn}. All experiments were performed on CPUs.


\subsection{Molecules}\label{app:exp_mol}

All models are parameterized with MPNN~\citep{gilmer2017neural}, for which we use the same paramaters as in previous works~\citep{bengio2021flow, malkin2022trajectory, madan2023learning}. The reward proxy and the test set are also taken from~\cite{bengio2021flow}. Following~\cite{madan2023learning}, we consider reward exponents from $\{4, 8, 10, 16\}$, learning rates from $\{ 5 \times 10^{-5}, 10^{-4}, 5 \times 10^{-4}, 10^{-3}\}$ and train all models for 50,000 iterations with a batch size of 4 using Adam optimizer. For \SubTB we use the same value $\lambda = 1.0$ as in~\cite{madan2023learning}. Reward correlations for the baselines in Figure~\ref{fig:mol_results} are taken directly from~\cite{madan2023learning}. We track Tanimoto-separated modes as described in~\cite{bengio2021flow}, using a raw reward threshold of $7.0$ and a Tanimoto similarity threshold of $0.7$.

Regarding \MDQN, we use Huber loss instead of MSE, following \cite{mnih2015human,vieillard2020munchausen} and find great benefits from utilizing dueling architecture \citep{wang2016dueling}, which builds similarity with \DB (see Section~\ref{sec:existing_gflow_as_softrl}). We also utilize prioritized replay buffer~\citep{schaul2016prioritized} from \texttt{TorchRL} library \citep{bou2023torchrl} of size $1,000,000$, and unusual parameters $\alpha=0.9, \beta=0.1$ (see Appendix~\ref{app:algo} for details on this choice), and use Munchausen parameters $\alpha=0.15$ and $l_0 = -2500$. At each training step, we sample 4 trajectories, put all transitions to the buffer, and then sample 256 transitions from the buffer to calculate the loss. To update the target network, we utilize soft updates akin to \texttt{DDPG} \citep{silver2014deterministic,lillicrap2015continuous} with a coefficient $\tau=0.1$. Notice that \MDQN samples exactly the same number of trajectories for training as the baselines, thus the comparison in terms of discovered modes is fair.

Our experiments are based upon the published code of~\cite{bengio2021flow} and~\cite{malkin2022trajectory}. We utilized a cluster with NVIDIA V100 and NVIDIA A100 GPUs for molecule generation experiments.

\subsection{Bit Sequences}\label{app:exp_bit}

Mode set $M$ and the test set are constructed as described in~\cite{malkin2022trajectory}. We also use the same Transformer~\citep{vaswani2017attention} neural network architecture for all models with 3 hidden layers, 64 hidden dimension and 8 attention heads, with the only difference from~\cite{malkin2022trajectory} that it outputs logits for a larger action space since we generate sequences non-autoregressively. 

To approximate $\cP_{\theta}(x)$ we use the Monte Carlo estimate proposed in~\cite{zhang2022generative}:
\begin{align*}
    \cP(x) = \mathbb{E}_{\PB(\tau \mid x)} \frac{\PF(\tau)}{\PB(\tau \mid x)} \approx \frac{1}{N} \sum\limits_{i = 1}^N \frac{\PF(\tau^i)}{\PB(\tau^i \mid x)},
\end{align*}
where $\PF(\tau) = \prod_{t=1}^{n_\tau} \PF\left(s_t | s_{t-1}, \theta\right)$, $\PB(\tau \mid x) = \prod_{t=1}^{n_\tau} \PB\left(s_{t-1} | s_{t}\right)$ and $\tau^i \sim \PB(\tau \mid x)$. \cite{zhang2022generative} showed that even with $N = 10$ this approach can provide an adequate estimate, so we use this value in our experiments.

We train all models for 50,000 iterations with a batch size 16 using Adam optimizer. For all methods the trajectories are sampled using a mixture of $\PF$ and a uniform distribution over next possible actions, the latter having a weight of $10^{-3}$ ($\varepsilon$-greedy exploration). 
We set the reward exponent value to $2$. For \MDQN and all baselines we pick the best learning rate from $\{ 5 \times 10^{-4}, 10^{-3}, 2 \times 10^{-3}\}$. For \SubTB we pick the best $\lambda$ from $\{0.9, 1.1, 1.5, 1.9\}$. 

For \MDQN we use Huber loss instead of MSE, following \citep{mnih2015human,vieillard2020munchausen} and priortized replay buffer \citep{schaul2016prioritized} from \texttt{TorchRL} library \citep{bou2023torchrl} of size $100,000$, and constants $\alpha=0.9, \beta=0.1$. At each training step, we sample 16 trajectories and put all transitions to the buffer, and then sample 256 transitions from the buffer to calculate the loss. Regarding specific Munchausen parameters \citep{vieillard2020munchausen}, we use $\alpha=0.15$ and $l_0 = -25$. We only use hard updates for the target network~\citep{mnih2015human} with a frequency of $5$ iterations. Notice that \MDQN samples exactly the same number of trajectories for training as the baselines, thus the comparison in terms of discovered modes is fair. Additionally, for \MDQN we found it helpful to increase the weight of the loss for edges leading to terminal states, which was also done in the original GFlowNet paper~\citep{bengio2021flow} for flow matching objective, and pick the best weight coefficient from $\{2, 5\}$ in our experiments.


For bit sequence experiments we used our own code implementation in PyTorch~\citep{paszke2019pytorch}. We also utilized a cluster with NVIDIA V100 and NVIDIA A100 GPUs.
\section{ADDITIONAL COMPARISONS}\label{app:ablation}
\vspace{-0.1cm}

In this section we provide additional experimental comparisons and ablation studies.

\vspace{-0.1cm}
\subsection{Hypergrids with Hard Reward}
\begin{figure*}[t!]

    \centering
    \includegraphics[width=0.276\linewidth]{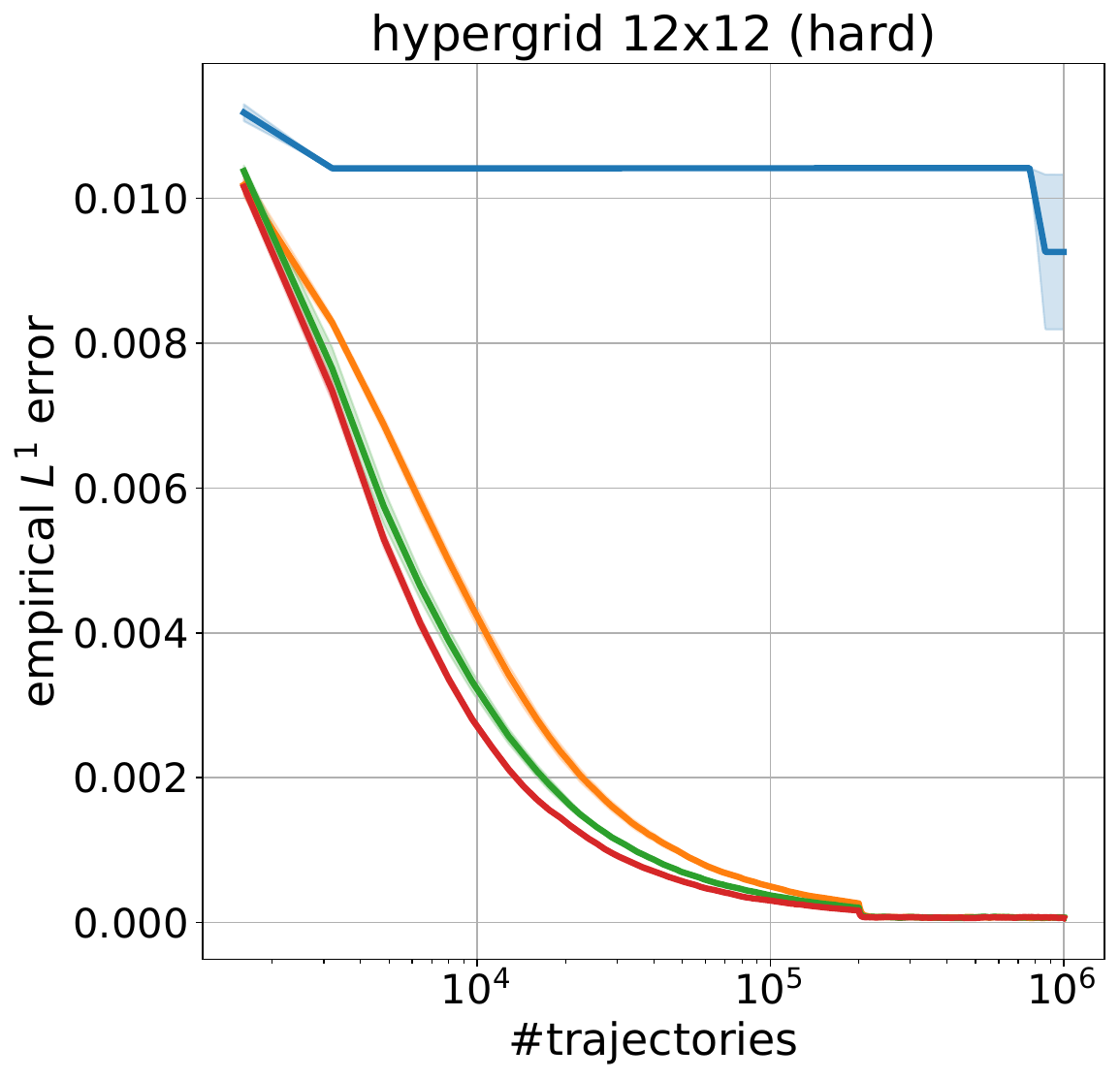}
    \includegraphics[width=0.27\linewidth]{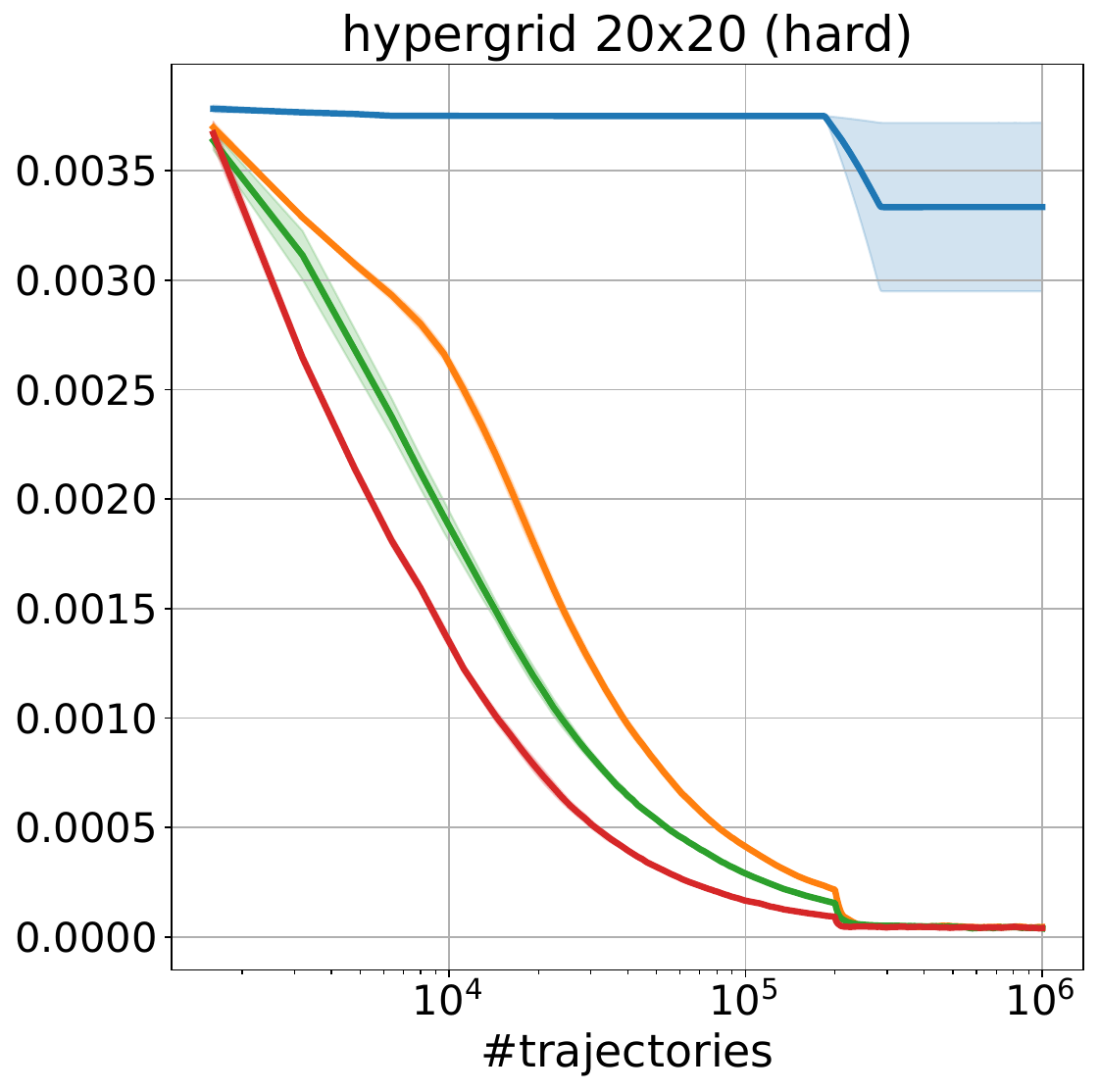}
    \includegraphics[width=0.27\linewidth]{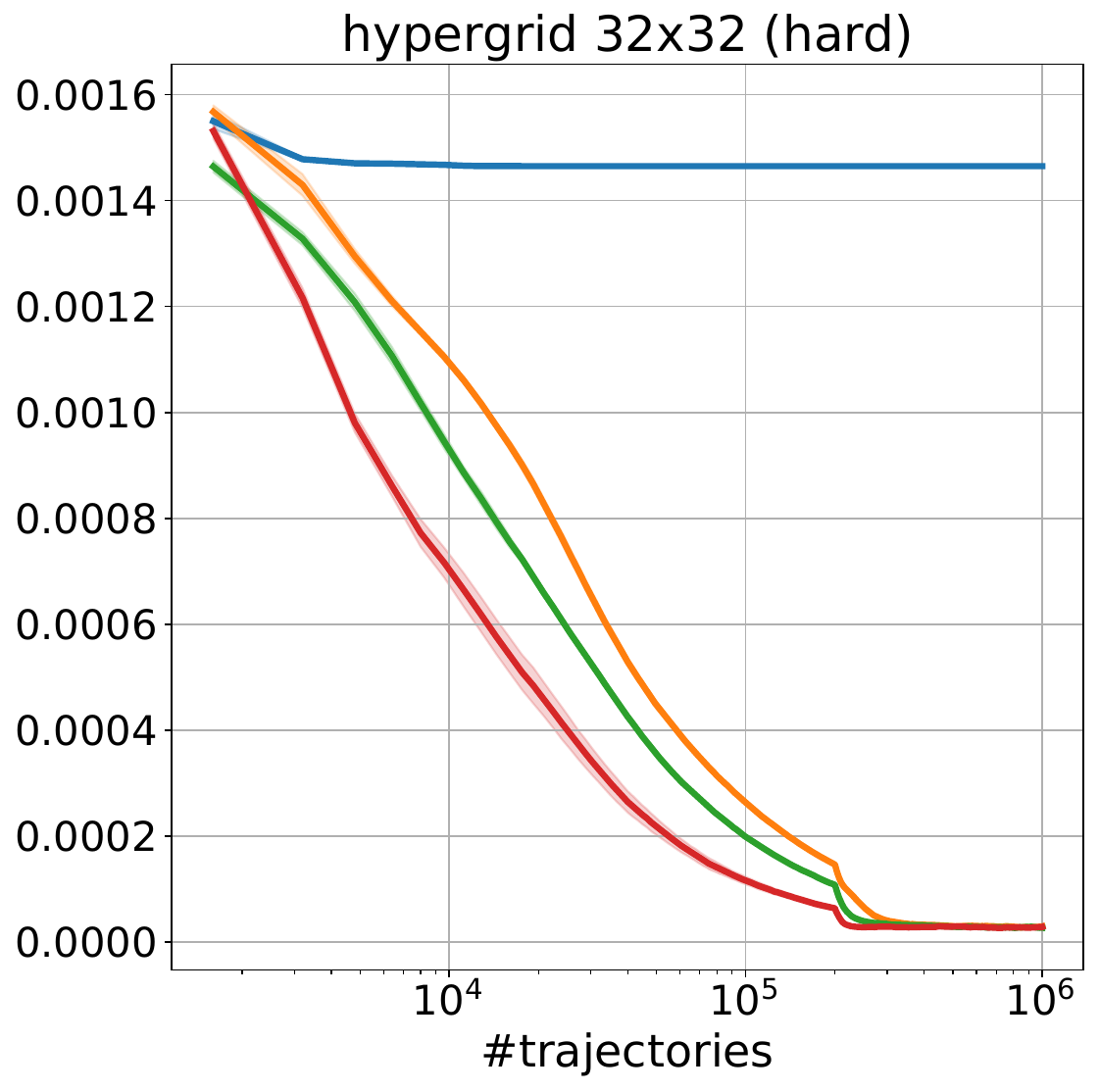}
    \includegraphics[width=0.276\linewidth]{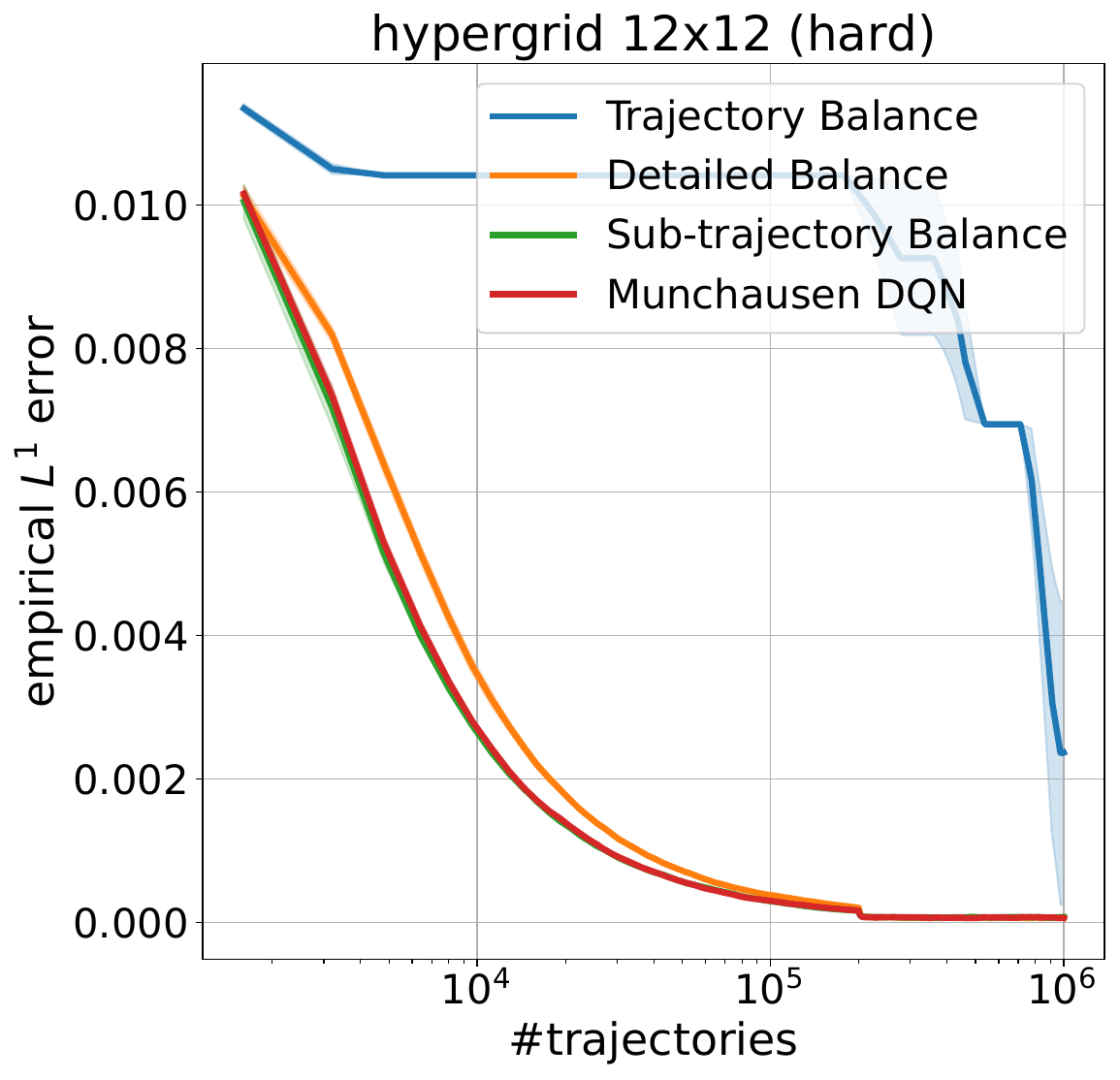}
    \includegraphics[width=0.27\linewidth]{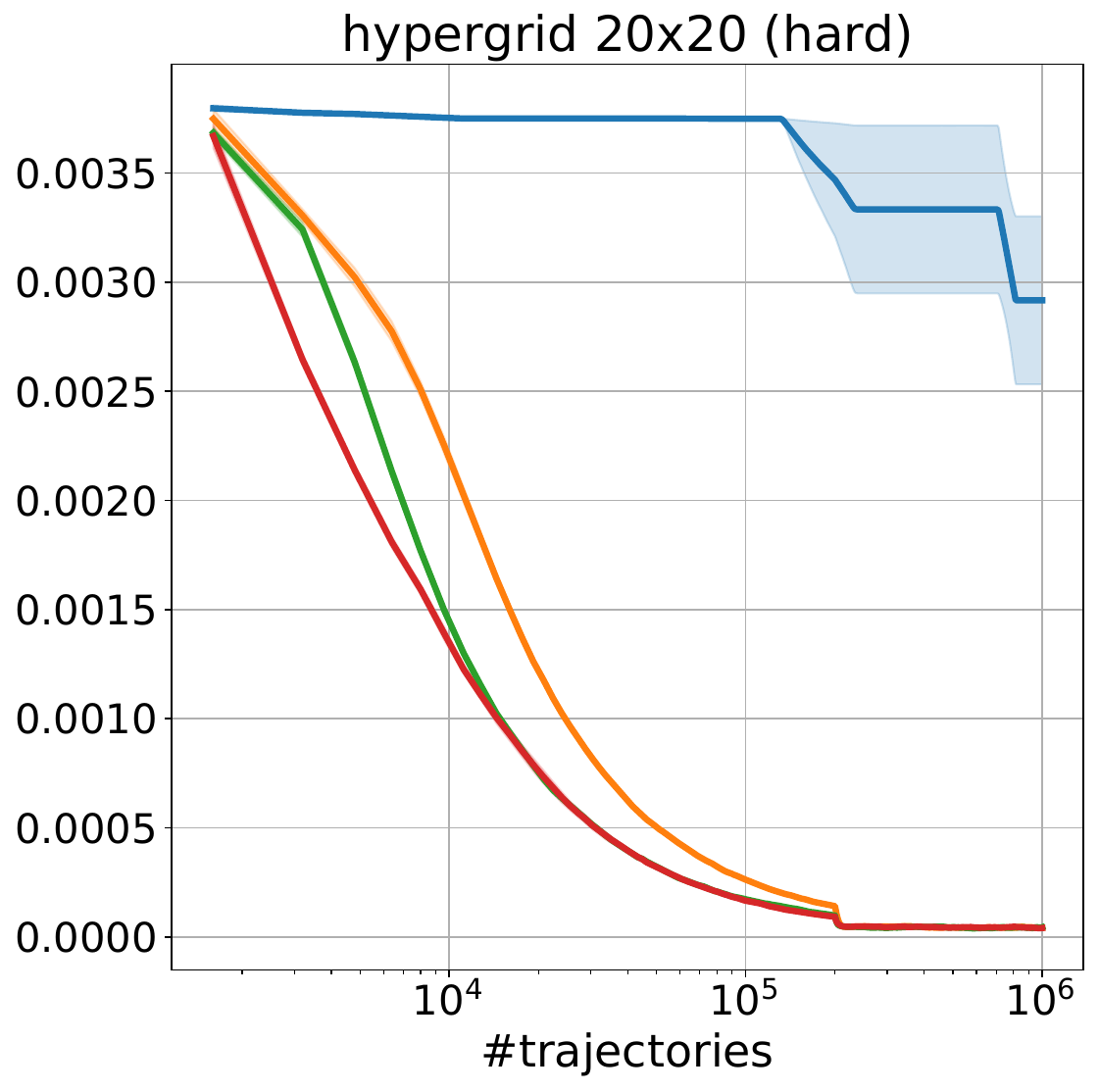}
    \includegraphics[width=0.27\linewidth]{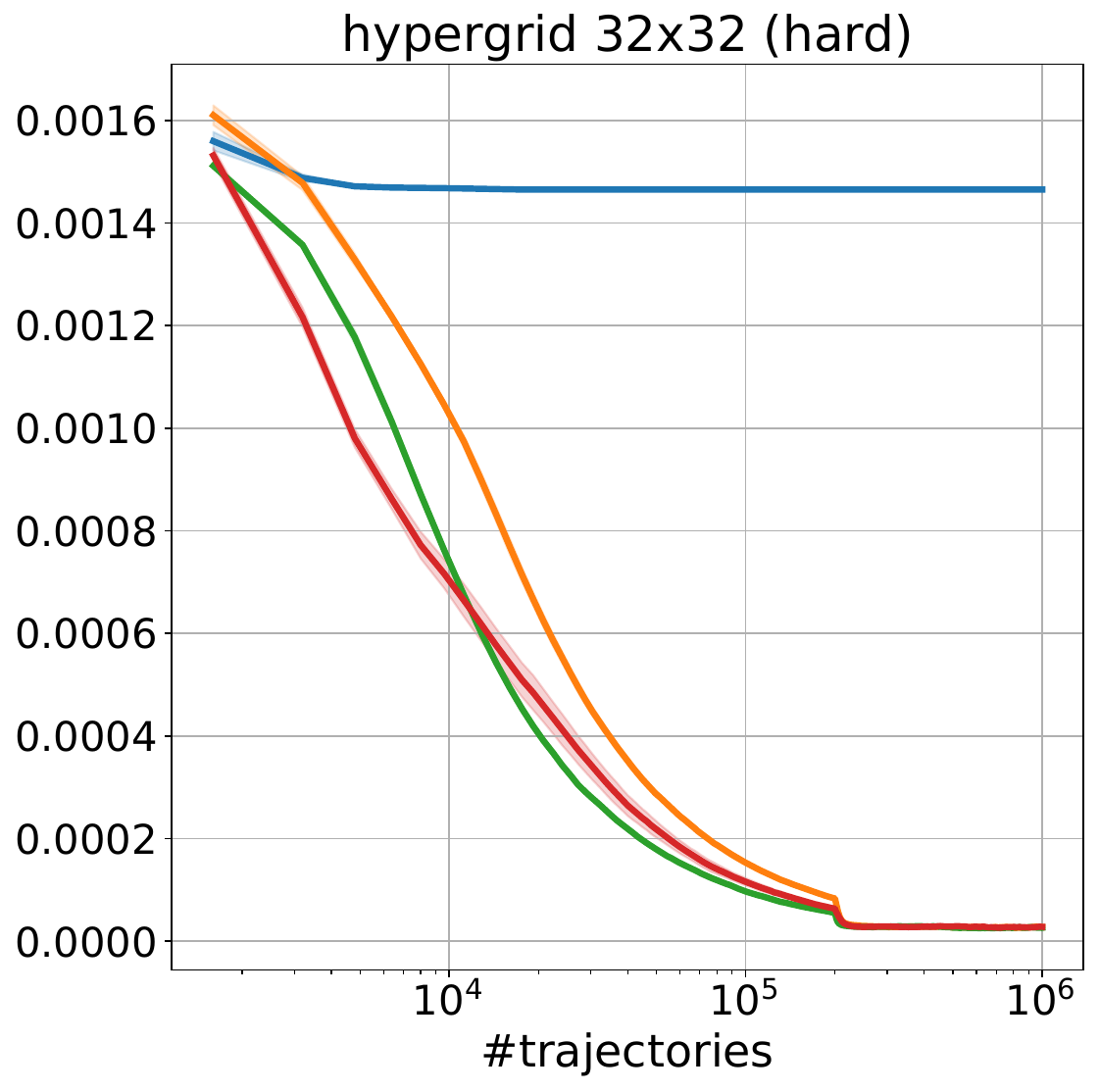}

    \caption{$L^1$ distance between target and empirical GFlowNet distributions over the course of training on the hypergrid environment for the hard reward variant. \textit{Top row:} $\PB$ is fixed to be uniform for all methods. \textit{Bottom row:} $\PB$ is learnt for the baselines and fixed to be uniform for \MDQN.} 
    \label{fig:hypergrid_hard}
    
\end{figure*}

\vspace{-0.1cm}
We provide experimental results for hypergrids with a harder reward variant proposed in~\cite{madan2023learning}, which has parameters $(R_0 = 10^{-4}, R_1 = 1.0, R_2 = 3.0)$. We follow the same experimental setup as in Section~\ref{sec:experiments}. The results are presented in Figure~\ref{fig:hypergrid_hard}. We make similar observations to the ones described in Section~\ref{sec:experiments}: when $\PB$ is fixed to be uniform for all methods, \MDQN converges faster than the baselines, and when $\PB$ is learnt for the baselines, \MDQN shows comparable performance to \SubTB and outperforms \TB and \DB. Notably, \TB fails to converge in most cases, which coincides with the results presented in~\cite{madan2023learning}.

\begin{figure*}[t!]

    \centering
    \includegraphics[width=0.275\linewidth]{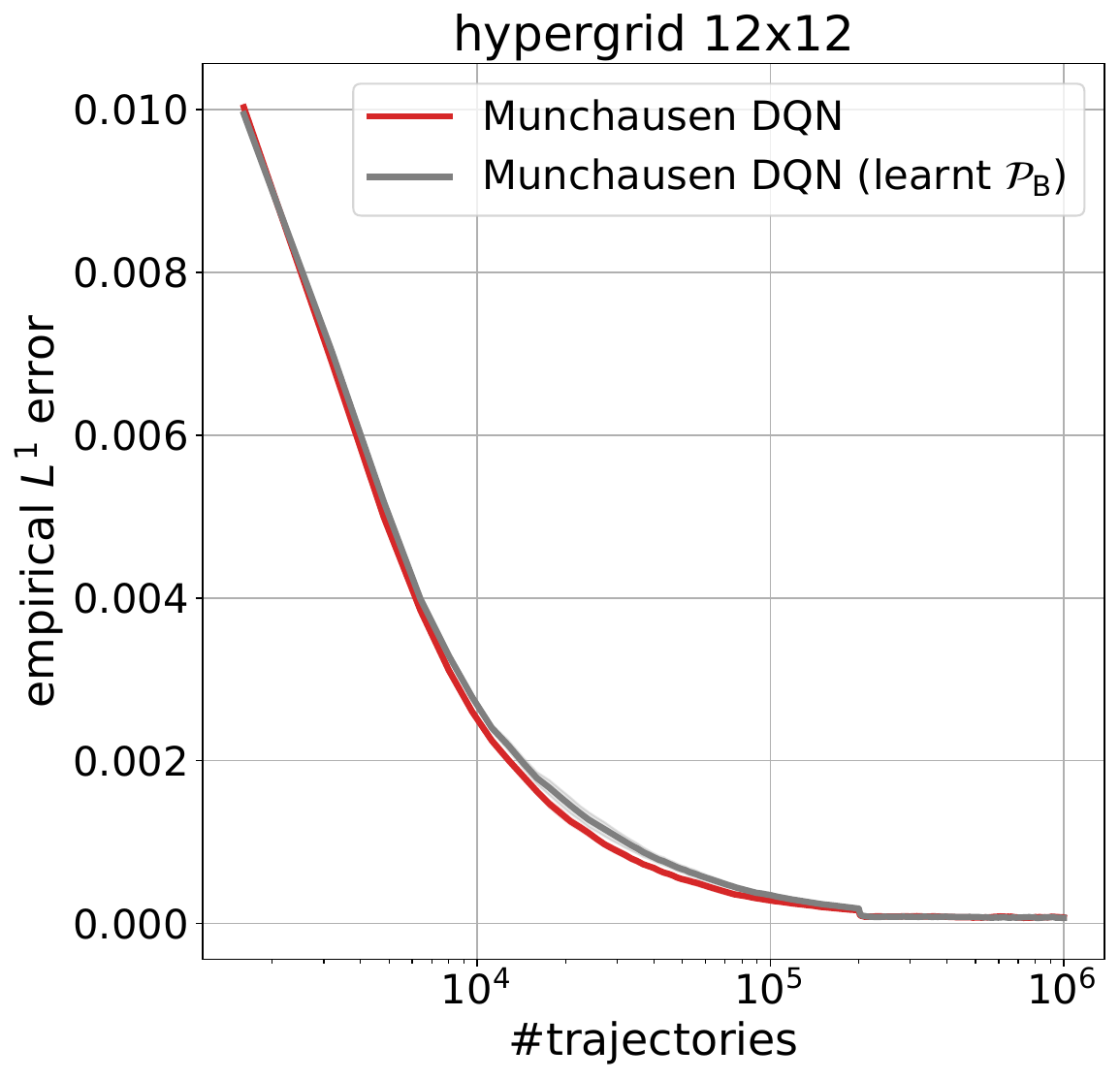}
    \includegraphics[width=0.27\linewidth]{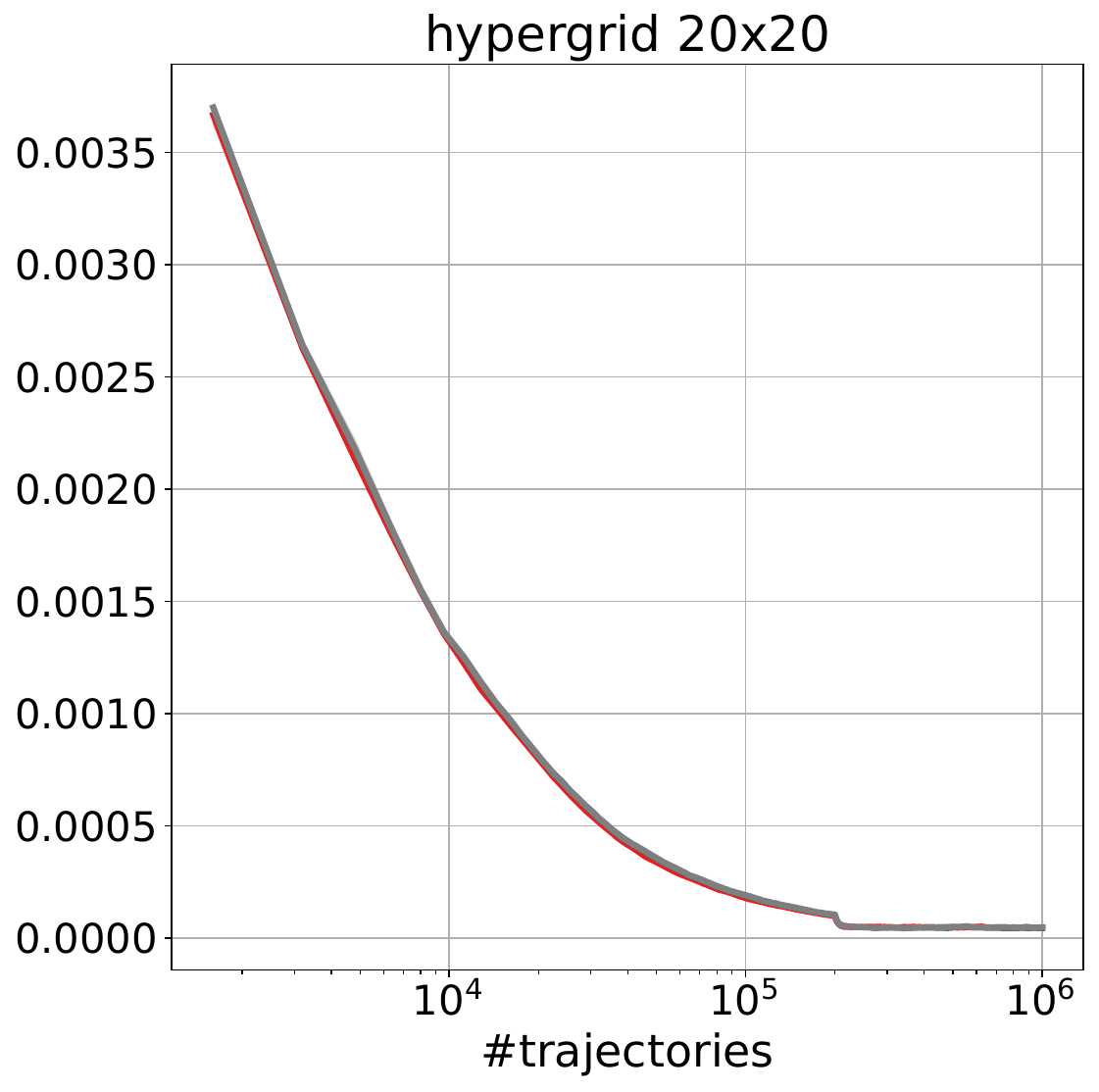}
    \includegraphics[width=0.27\linewidth]{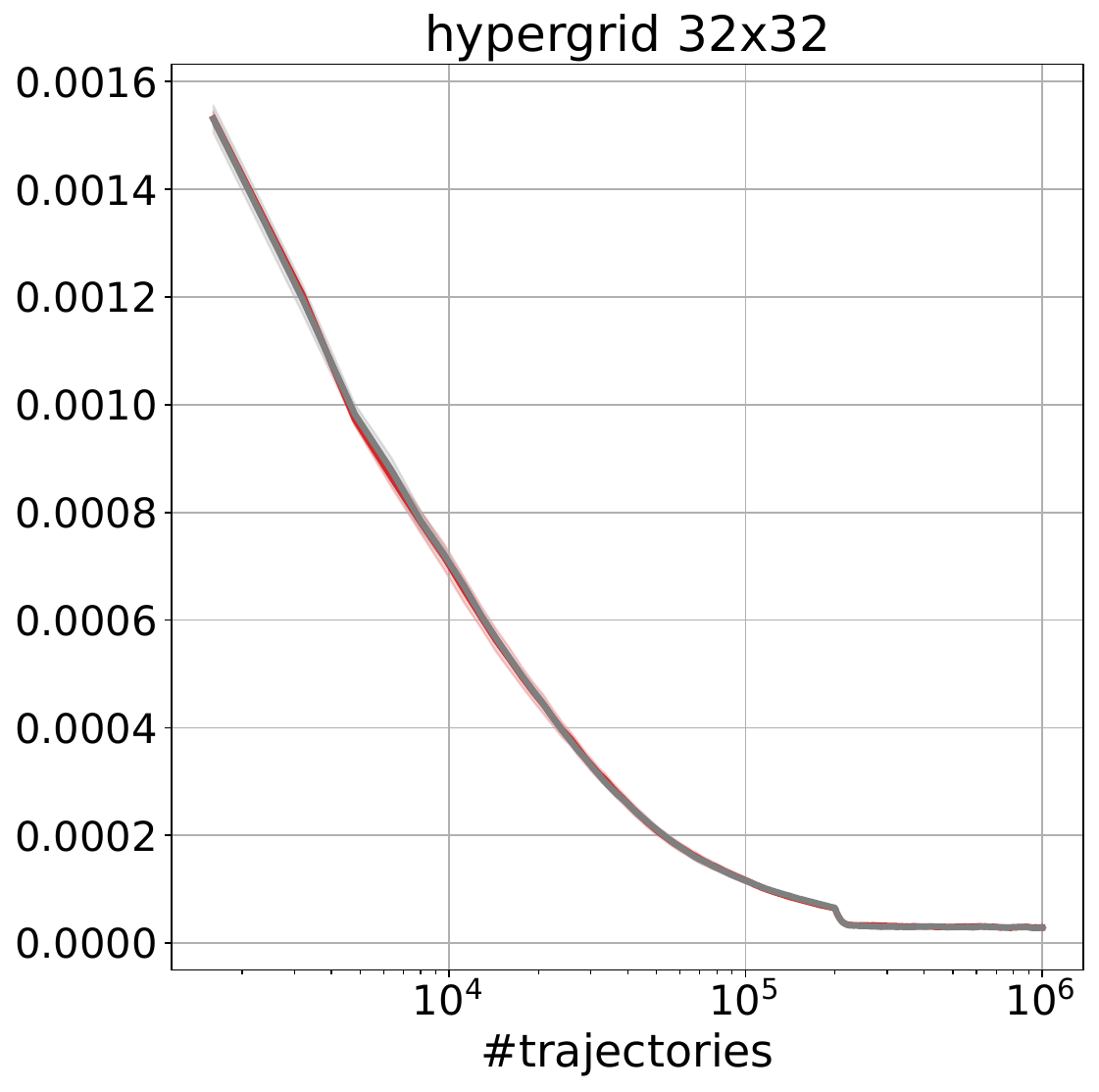}
    \includegraphics[width=0.254\linewidth]{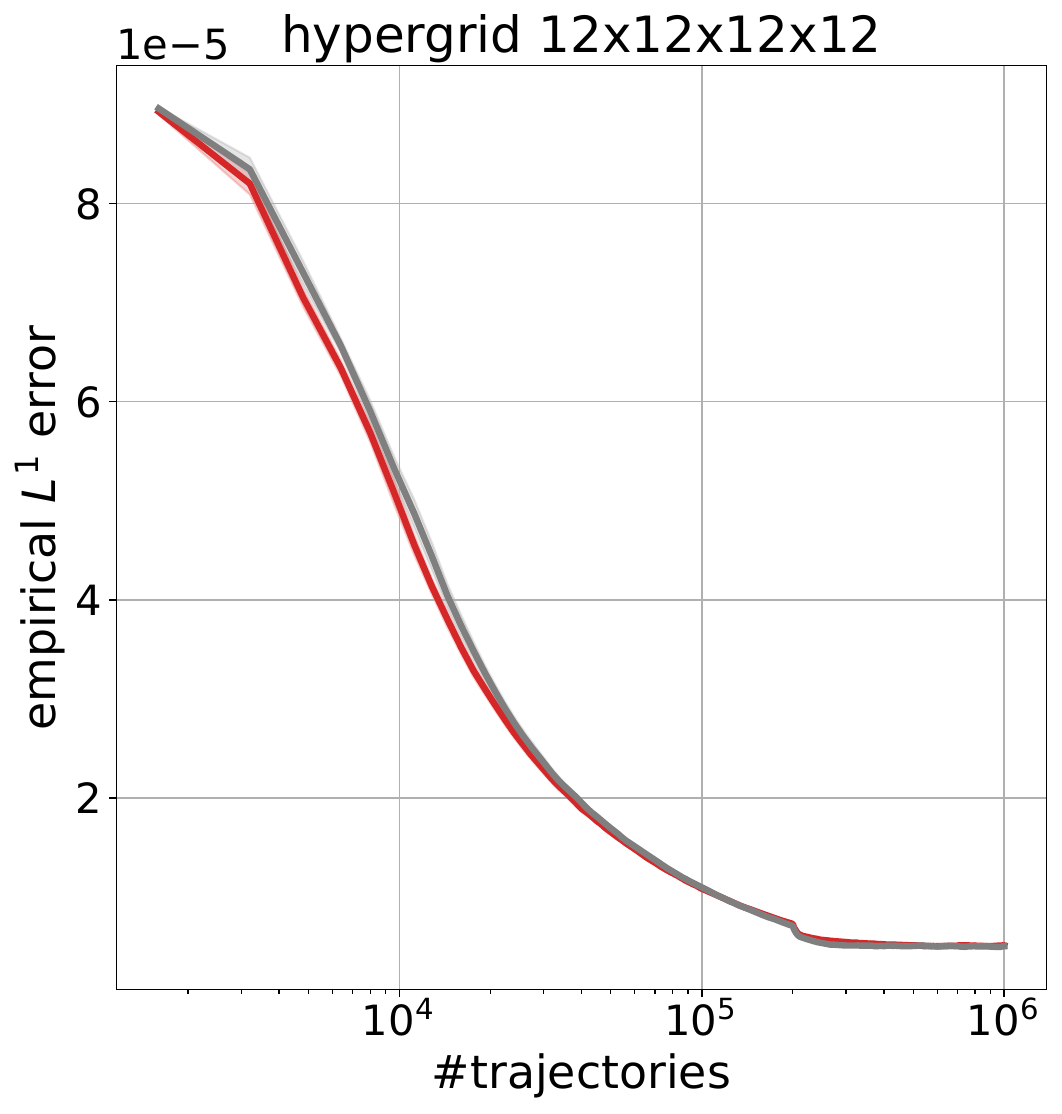}
    \;
    \includegraphics[width=0.250\linewidth]{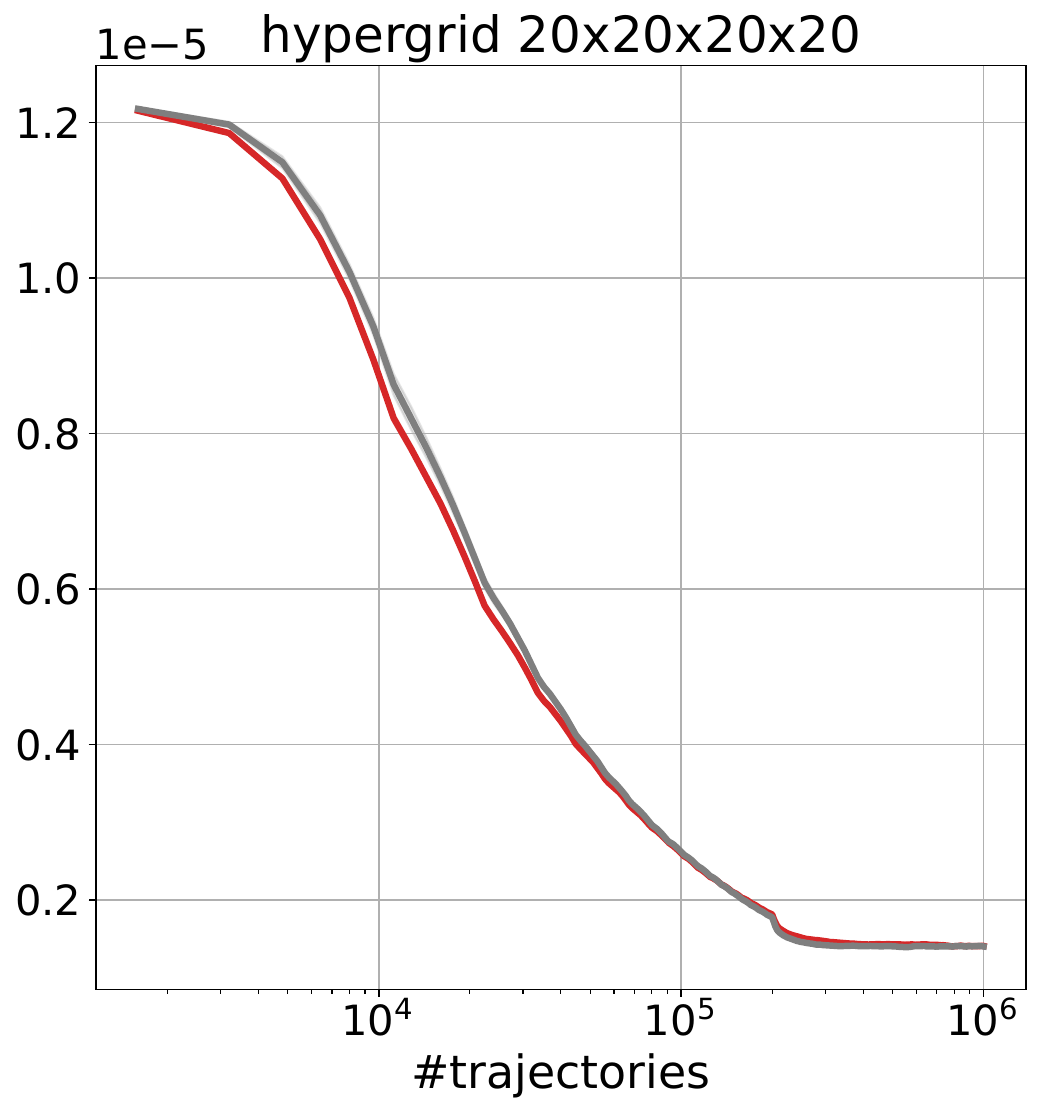}
    \;
    \includegraphics[width=0.250\linewidth]{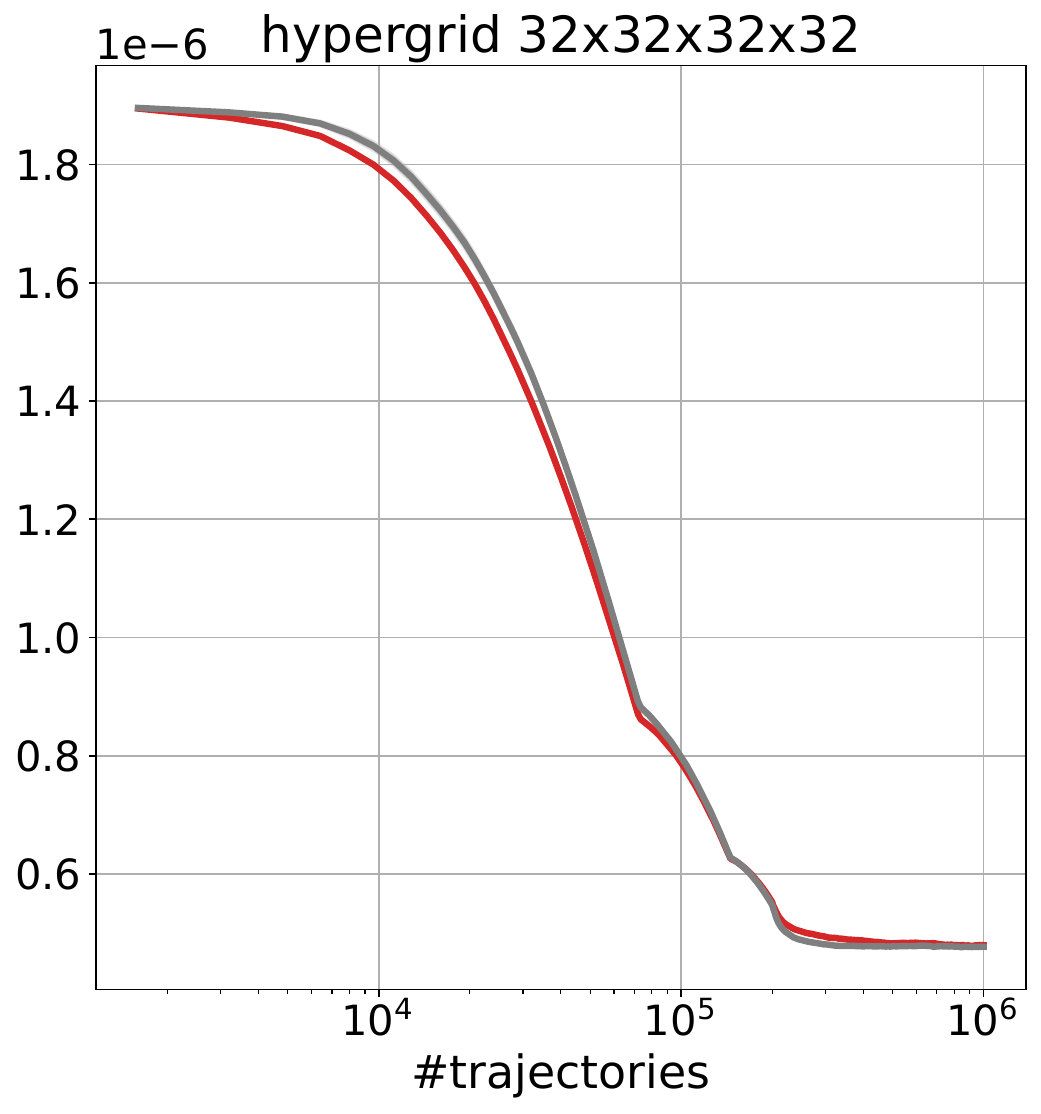}
    
    \caption{$L^1$ distance between target and empirical GFlowNet distributions over the course of training on the hypergrid environment. Here we present two versions of \MDQN: one with a fixed uniform $\PB$ and another one with a learnt $\PB$ during the training.} 
    \label{fig:hypergrid_lpb}
\end{figure*}

\subsection{Training Backward Policy}
\vspace{-0.1cm}

Our theoretical framework does not support running soft RL with a learnable backward policy, thus in Section~\ref{sec:experiments} we used \MDQN with uniform $\PB$ in all cases. Nevertheless, one can try to use the same \MDQN training loss to calculate gradients and update $\PB$, thus creating a setup with a learnable backward policy. Figure~\ref{fig:hypergrid_lpb} presents the results of comparison on the hypergrid environment (with a standard reward, the same as in Section~\ref{sec:experiments}). We find that having a trainable $\PB$ does not improve the speed of convergence for \MDQN, in contrast to the behavior of a learnt $\PB$ in the case of the previously proposed GFlowNet training methods presented in Figure~\ref{fig:hypergrid_base}. 

Another possible approach can be to use some other GFlowNet training objective to calculate a loss for $\PB$, for example \TB, while using soft RL to train $\PF$. We leave this case for further study. 


\subsection{Different Soft RL algorithms}

\begin{figure*}[t!]

    \centering
    \includegraphics[width=0.275\linewidth]{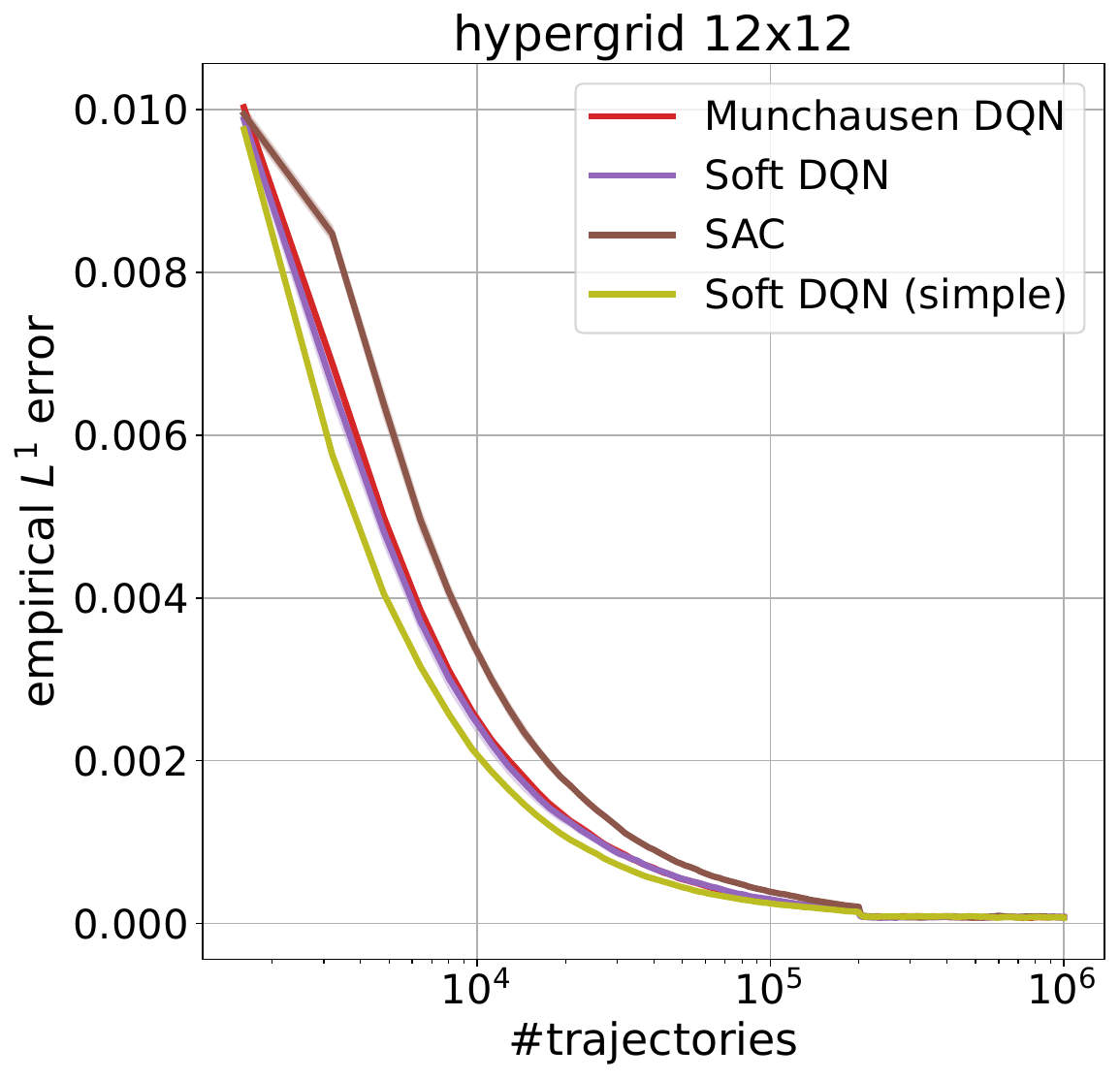}
    \includegraphics[width=0.27\linewidth]{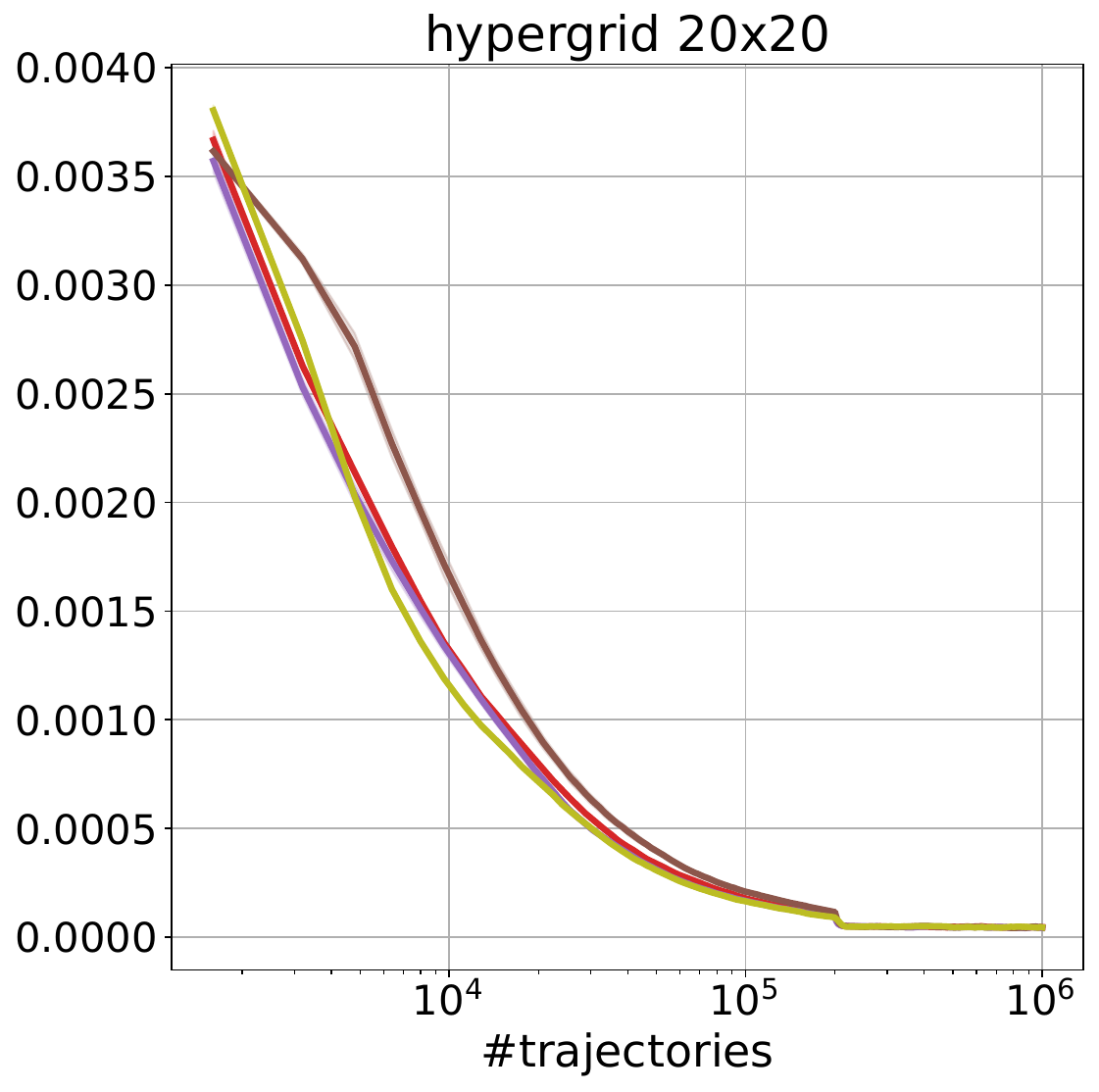}
    \includegraphics[width=0.27\linewidth]{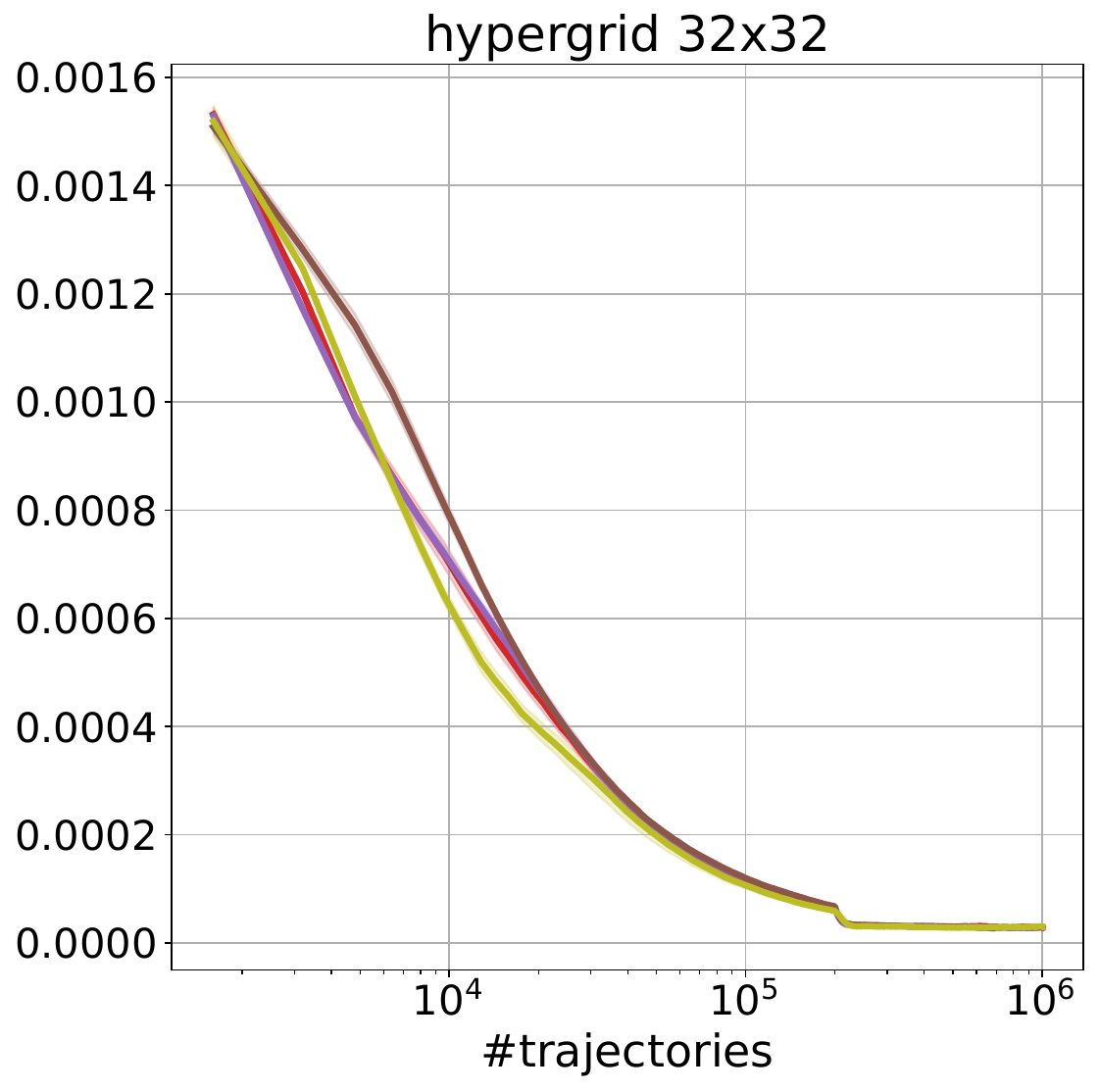}
    \includegraphics[width=0.254\linewidth]{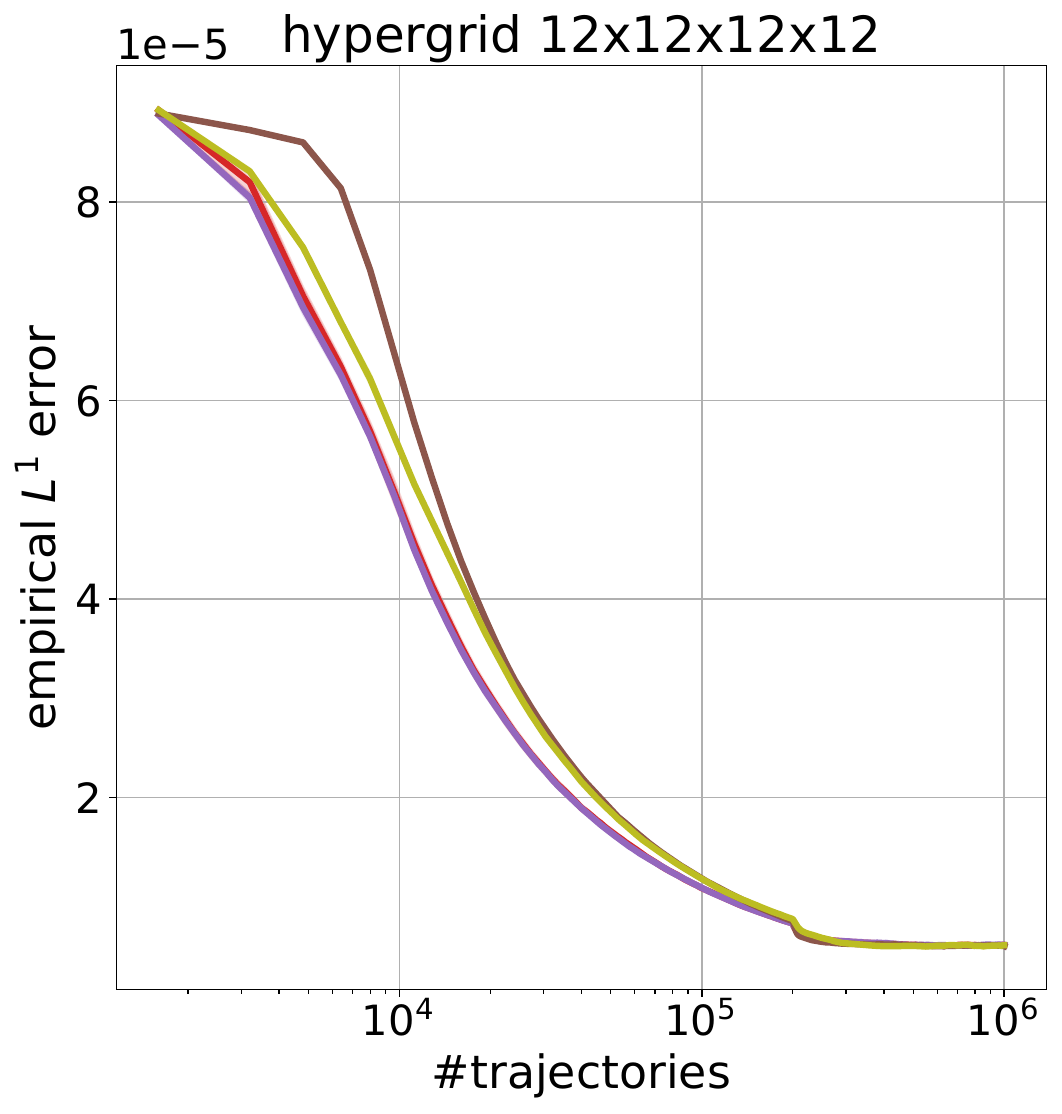}
    \;
    \includegraphics[width=0.250\linewidth]{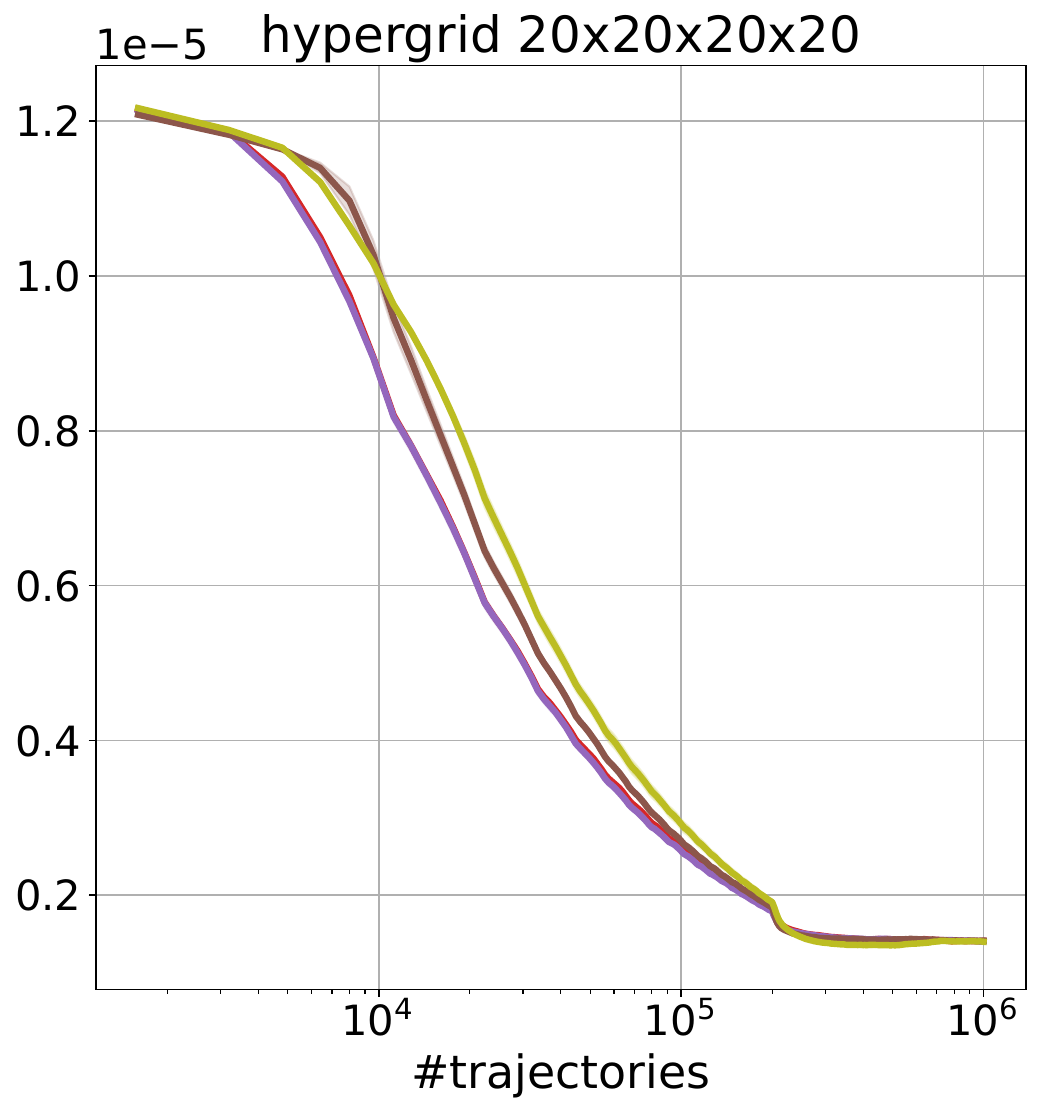}
    \;
    \includegraphics[width=0.250\linewidth]{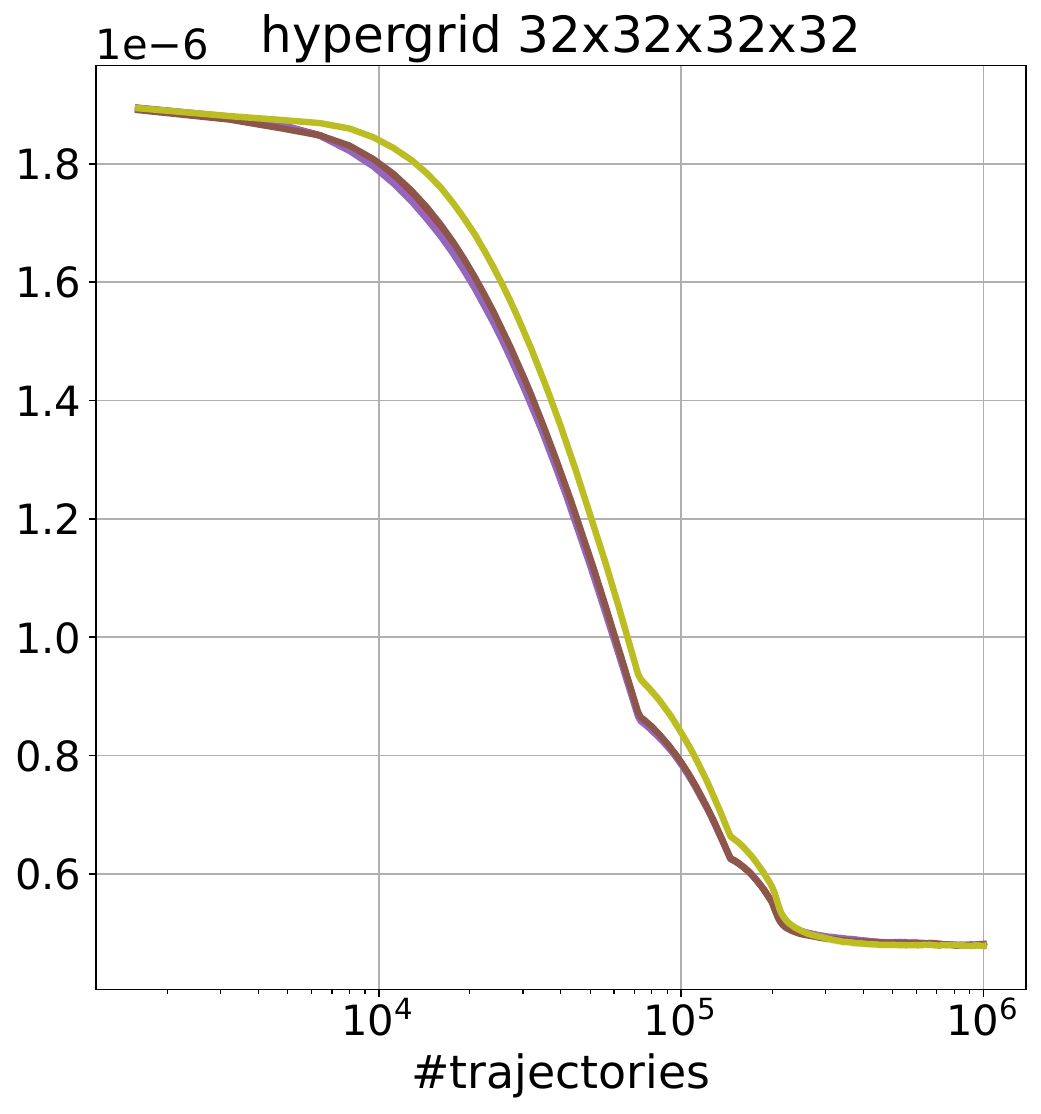}
    
    \caption{$L^1$ distance between target and empirical GFlowNet distributions over the course of training on the hypergrid environment. $\PB$ is fixed to be uniform for all methods. Here we compare different soft RL algorithms. 
    }
    \label{fig:hypergrid_other_rl}
    
\end{figure*}

\subsubsection{Hypergrid Environment}

We also compare \MDQN to other known off-policy soft RL algorithms: 

\paragraph{Soft DQN}

To ablate the effect of Munchausen penalty in \MDQN we compare it to a classical version of \SoftDQN, which is equivalent to setting the Munchausen constant $\alpha = 0$. All other hyperparameters are chosen exactly the same as for \MDQN (see Appendix~\ref{app:exp}). 

\paragraph{Soft DQN (simple)}
To ablate the effect of using a replay buffer and Huber loss, we also train a simplified version of \SoftDQN. At each training iteration, it samples a batch of trajectories and uses the sampled transitions to calculate the loss, similarly to the way GFlowNet is trained with \DB objective. It also uses MSE loss instead of Huber.

\paragraph{Soft Actor Critic}
We also considered \SAC \citep{haarnoja2018soft} algorithm, namely its discrete counterpart presented by \cite{christodoulou2019soft}. We follow the implementation presented in \texttt{CleanRL} library \citep{huang2022cleanrl} with two differences: usage of Huber loss \citep{mnih2015human} and lack of entropy regularization parameter tuning (since our theory prescribes to set its value to $1$). All other hyperparameters are chosen exactly the same as for \MDQN and \SoftDQN (see Appendix~\ref{app:exp}).  

Figure~\ref{fig:hypergrid_other_rl} presents the results of comparison on the hypergrid environment (with standard reward, the same as in Section~\ref{sec:experiments}). We found that \MDQN and \SoftDQN generally have very similar performance, while converging faster than \SAC in most cases. Interestingly, the simplified version of \SoftDQN holds its own against other algorithms, converging faster on smaller 2-dimensional grids while falling behind \MDQN and \SoftDQN on 4-dimensional ones. This indicates that the success of soft RL is not only due to using replay buffer and Huber loss instead of MSE, but the training objective itself is well-suited for the GFlowNet learning problem.

\subsubsection{Small Molecule Generation}

\begin{figure*}[h!]

  \centering
  
\begin{tabular}{@{}c@{}}
    \includegraphics[width=0.47\linewidth]{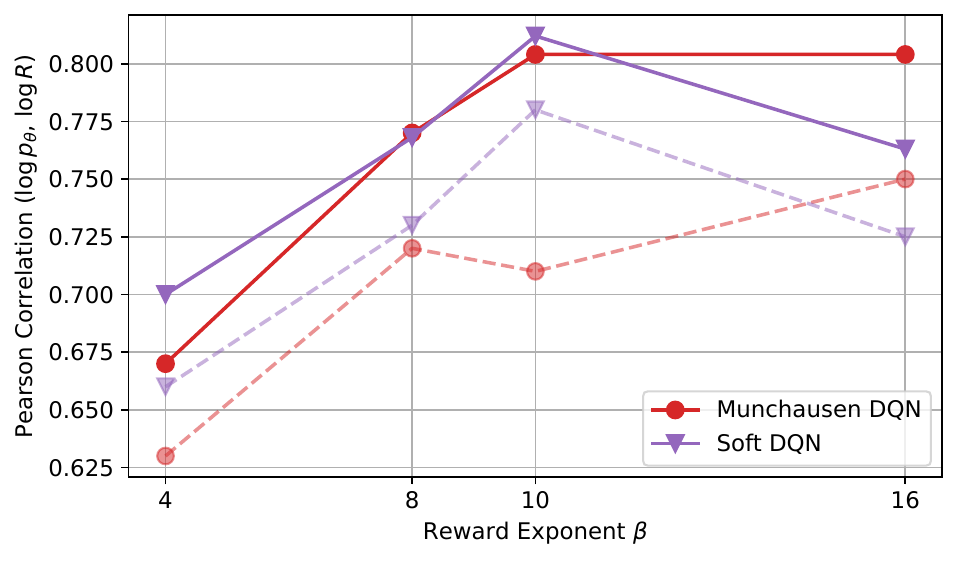} 
\end{tabular}
  \begin{tabular}{@{}c@{}}
    \includegraphics[width=0.47\linewidth]{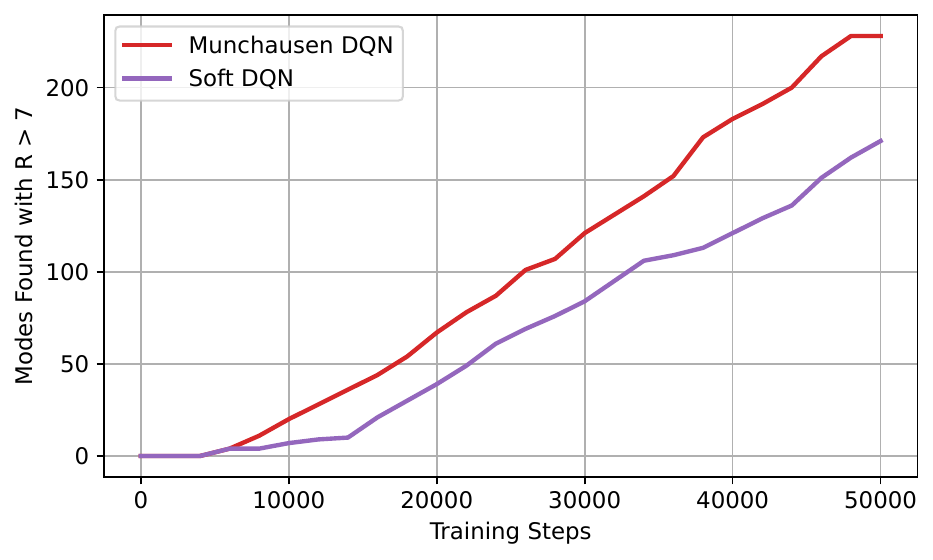} 
\end{tabular}
  \caption{
Small molecule generation results. \textit{Left:} Pearson correlation between $\log \cR$ and $\log \cP_{\theta}$ on a test set for each method and varying $\beta \in \{4, 8, 10, 16\}$. Solid lines represent the best results over choices of learning rate, dashed lines — mean results. \textit{Right:} Number of Tanimoto-separated modes with $\tilde{\cR} > 7.0$ found over the course of training for $\beta = 10$.}\label{fig:mol_results_soft}

\end{figure*}

We also compare \MDQN and \SoftDQN on the molecule generation task. The experimental setup is the same as the one used in Section~\ref{sec:experiments}. Figure~\ref{fig:mol_results_soft} presents the results. \SoftDQN shows worse performance when $\beta = 16$ in comparison to \MDQN. But in all other cases, it shows comparable or better reward correlations and is more robust to the choice of learning rate. However, \MDQN discovers modes faster.

\subsubsection{Bit Sequence Generation}

\begin{figure}[h!]

  \centering
  \begin{tabular}{@{}c@{}}
    \includegraphics[width=0.47\linewidth]{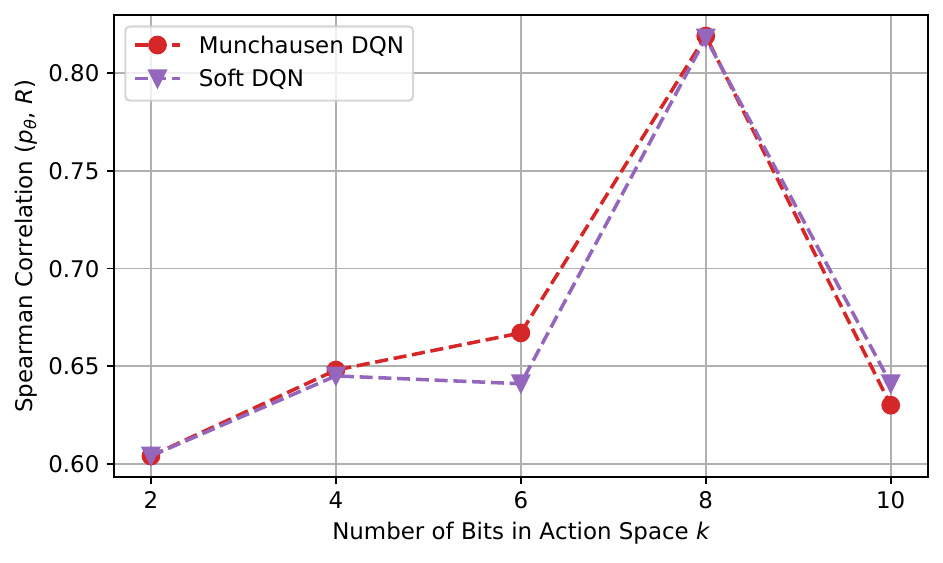} 
  \end{tabular}
  \begin{tabular}{@{}c@{}}
    \includegraphics[width=0.457\linewidth]{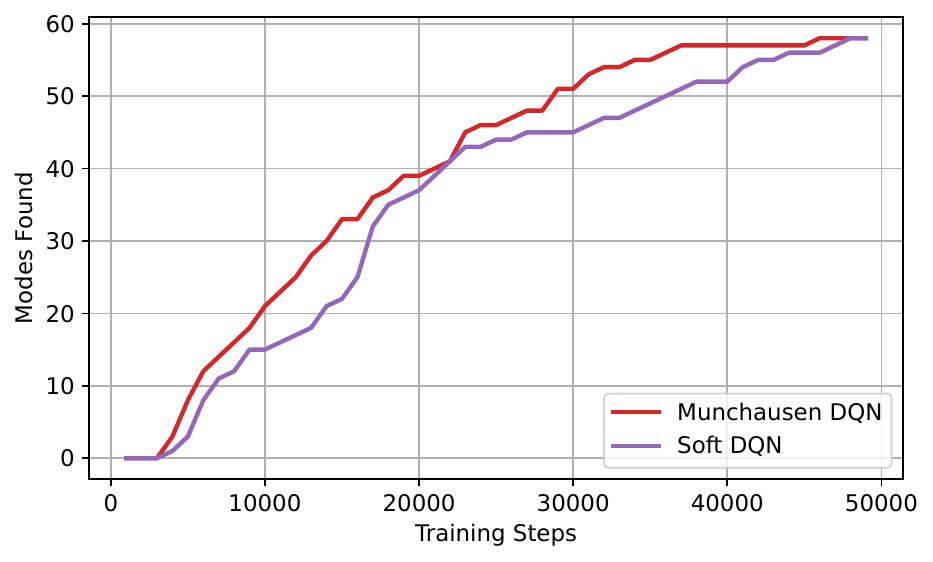} 
  \end{tabular}

  \caption{Bit sequence generation results. \textit{Left:} Spearman correlation between $\cR$ and $\cP_{\theta}$ on a test set for each method and varying $k \in \{2, 4, 6, 8, 10\}$. \textit{Right:} The number of modes discovered over the course of training for $k = 8$.}\label{fig:bit_results_soft}

\end{figure}

Finally, we compare \MDQN and \SoftDQN on the bit sequence generation task. The experimental setup is the same as the one used in Section~\ref{sec:experiments}. Figure~\ref{fig:bit_results_soft} presents the results. \SoftDQN has comparable or worse reward correlations in comparison to \MDQN and discovers modes slightly slower.

\vfill 

\end{document}